%% file: main.tex
\documentclass[letterpaper,journal]{IEEEtran}
\usepackage{amsmath,amsfonts,amssymb}
\usepackage{algorithmic}
\usepackage{algorithm}
\usepackage{array}
\usepackage{booktabs}
\usepackage{multirow}
\usepackage[caption=false,font=normalsize,labelfont=sf,textfont=sf]{subfig}
\usepackage{textcomp}
\usepackage{stfloats}
\usepackage{url}
\usepackage{verbatim}
\usepackage{graphicx}
\usepackage{cite}
\usepackage{xspace}
\usepackage{xcolor}
\usepackage{tikz}
\usetikzlibrary{arrows.meta,positioning,calc,backgrounds,fit,shadows.blur}
\newcommand{\method}{ConGraspXL\xspace}

\begin{document}

\title{ConGraspXL: Controllable Constraint-Conditioned Dexterous Grasping Motion Synthesis}

\author{Hui Zhang, Mirko Meboldt, Jie Song
        \thanks{Hui Zhang and Mirko Meboldt are with ETH Zürich, Switzerland
         }
         
        \thanks{Jie Song is with HKUST (GZ) and HKUST, China}
        \thanks{This work has been submitted to the IEEE for possible publication. Copyright may be transferred without notice, after which this version may no longer be accessible.}
        }


\maketitle

\input{sec/00_abstract}

\input{sec/01_introduction}

\input{sec/02_related_work}

\input{sec/03_method}

\input{sec/04_experiment}

\input{sec/10_conclusion}

\bibliographystyle{IEEEtran}
\bibliography{egbib}

\end{document}

%% file: sec/00_abstract.tex
\begin{abstract}
Dexterous grasping is usually conducted for specific tasks, leading to heterogeneous constraints such as specific approach directions, desired contact regions, specified wrist trajectories, and functional hand poses. Our previous work, GraspXL, achieves scalable grasping motion synthesis for diverse objects and hand morphologies, while lacking controllability for synthesis under such various task-driven constraints.
In this paper, we propose \method, which extends GraspXL with controllable constraint-conditioned grasp motion synthesis that accommodates diverse task-driven constraints and their combinations.
We introduce a hierarchical constraint formulation, enable flexible constraint composition with a masked residual interface, and improve control precision with dynamic hand centers
and feed-forward wrist guidance. Without losing the strong generalization capabilities of GraspXL, \method enables precise and flexible controllability for various individual constraints and their combinations, providing a plug-and-play low-level grasp controller for downstream applications such as whole-body grasp completion, functional grasping, and human-motion imitation.

\end{abstract}


%% file: sec/01_introduction.tex
\section{Introduction}

\input{figures/teaser}

The remarkable dexterity of the human hands enables their interaction with different objects in diverse ways for various tasks.
Among all the interactions, grasping is one of the most fundamental and frequent skills.
Instead of grasping objects arbitrarily, we typically grasp objects following specific requirements imposed by tasks.
For example, cutting with a knife requires holding the handle, using scissors calls for a functional grasping pose, and grasping in clutter may require a particular approach direction to avoid colliding with other objects.
The ability to generate grasping motions following such task-driven constraints is thus valuable for a wide range of downstream applications, including virtual reality~\cite{Han2025ForceGrip, han2024vrhandnet}, animation~\cite{ye2012synthesis}, and robotics~\cite{zhang2025RobustDexGrasp}.

However, dexterous grasping synthesis is inherently difficult due to the complex dynamic interactions and diverse object geometries, which is made even harder by the varied constraints imposed by different tasks. As a result, most prior work still focuses on producing static grasp configurations through optimization~\cite{grady2021contactopt, Brahmbhatt_2020_ECCV, liu2023contactgen} or data-driven methods~\cite{grab, karunratanakul2020grasping, jiang2021graspTTA}, while ignoring the temporal dynamic interactions. Some recent works synthesize temporal grasping motions~\cite{taheri2021goal, christen2024diffh2o, ghosh2022imos}, yet they generalize poorly to out-of-distribution objects and hand morphologies. Moreover, they rarely address controllability for various task-driven motion constraints.

Our prior work, GraspXL~\cite{zhang2024graspxl}, takes an important step toward generalizable grasping motion generation. It shows that physics-based reinforcement learning can scale dexterous grasp motion generation to 500k+ unseen object geometries and diverse hand morphologies without relying on any hand-object data. Its policies and generated interaction data have since supported a broad range of downstream applications for the community, such as HOI video generation~\cite{gao2025PAM}, 3D hand-object reconstruction~\cite{liu2026hggt,chen2026forehoifeedforward3dobject}, large-scale hand pose estimation~\cite{si2026anyhand}, and dexterous robot grasping~\cite{bortolon2025grasplat,park2026demodiffusion}.
However, GraspXL is designed primarily for generalization, and its controllability is tied to a fixed and limited set of objectives, namely a wrist-centric specification (heading direction, position, and wrist rotation) together with an affordance region.
This narrow and rigid conditioning interface prevents GraspXL from serving as a universal low-level grasping motion generator under heterogeneous constraints that upstream tasks actually produce, such as full wrist trajectories from a whole-body planner or static functional reference poses retargeted from human demonstrations.

In this paper, we focus on controllable dexterous grasp motion synthesis conditioned on various task-driven constraints.
While preserving the strong generalization of GraspXL, we systematically define, formulate, and evaluate a broad family of constraints, leading to a plug-and-play
grasp motion generator for varied constraints imposed by different tasks.
Achieving this within a single policy poses three challenges.
First, different tasks impose heterogeneous constraints, ranging from a coarse approaching direction to a dense wrist trajectory or a per-joint reference pose, which are hard to model within a unified interface.
Second, constraints rarely appear in a fixed configuration. They may be specified individually or jointly and present or absent for a given task, which requires high flexibility of the policy.
Finally, controllability must be achieved without sacrificing generalization. Faithfully satisfying constraints while keeping grasps physically stable across diverse objects and hand morphologies demands higher control precision, especially on challenging object shapes.

To meet these challenges, we cast controllable constraint-conditioned dexterous grasping motion synthesis as a reinforcement learning problem trained in a physics simulator.
To unify heterogeneous constraints, we organize them into four semantic levels: heading-level (desired hand heading direction and position), pose-level (desired final finger poses and wrist pose), trajectory-level (desired wrist motions), and affordance-level (desired contact area), based on which a priority-aware composition rule is proposed to resolve overlaps.
To flexibly support optional and combinable constraints, we inject each constraint as a masked residual feature with an explicit validity mask across the observation and reward spaces, and randomize the activated constraints during training.
To facilitate control precision, we propose dynamic hand centers for adaptative motion generation to the object sizes
and extend the hand guidance of GraspXL to a feed-forward wrist guidance term for faithful trajectory tracking.

We conduct systematic evaluations on controllability, generalization, 
and downstream usability. Our method enables flexible controllability under heterogeneous task constraints, with some examples shown in Figure~\ref{fig:teaser}. It accurately satisfies individual and compositional constraints while maintaining high grasp success rates, outperforming specialized baselines designed for individual constraints. At the same time, it preserves the strong cross-object and cross-hand generalization capabilities of GraspXL for controllable synthesis.
We also demonstrate that our framework serves as a plug-and-play low-level grasp controller for a variety of applications, including completing hand motions from whole-body wrist trajectories, generating functional grasps from reference poses, and imitating human hand motions with dexterous robot hands.

The contributions of this paper are summarized as follows:
\begin{itemize}
    \item We propose \method, a dexterous grasping motion synthesis method that transforms heterogeneous task-driven constraints into controllable grasp behaviors.
    \item We formulate task-driven constraints in a hierarchical structure with four semantic levels and a masked residual conditioning, leading to a unified constraint interface that can flexibly support optional, combinable, and overlapping constraints.
    \item We introduce dynamic hand centers
    and feed-forward wrist guidance, which effectively improve the stability and accuracy of constrained dexterous grasping.
    \item We conduct comprehensive evaluations showing that \method achieves strong controllability while preserving generalization, 
    and can serve as a plug-and-play motion generator for various downstream applications.
\end{itemize}

\noindent\textbf{Relation to GraspXL.}
This paper is an extension of our previous work GraspXL~\cite{zhang2024graspxl}, a conference paper published at ECCV'24 which scales physics-based dexterous grasping motion generation to diverse objects and hand morphologies. However, GraspXL only exposes controllability on limited constraint modalities.
This paper extends GraspXL with a specific focus on the controllability of constraint-conditioned grasping motion generation, while keeping the strong generalization capabilities. We introduce a set of heterogeneous constraints with a hierarchical constraint formulation, enable flexible constraint composition with masked residual interface, and improve control precision with techniques including 
dynamic hand centers and feed-forward wrist guidance.
We also provide a comprehensive evaluation which verifies the controllability
without generalization degradation, and demonstrates the downstream usability of the proposed method.

%% file: figures/teaser.tex
\newcommand{\teaserrefbox}[2]{%
    \begin{tikzpicture}
        \node[draw=black!30, line width=0.4pt, rounded corners=2pt,
              inner sep=3pt] {%
            \begin{minipage}{\dimexpr\linewidth-8pt\relax}
                \centering
                {\scriptsize\sffamily\strut #1}\\[2pt]
                \parbox[b][0.883\linewidth][c]{\linewidth}{\centering #2}%
            \end{minipage}};
    \end{tikzpicture}%
}

\newcommand{\teasermotion}[1]{%
    \parbox[b][0.879\linewidth][c]{\linewidth}{\centering #1}%
}

\newlength{\teasercolw}
\newlength{\teaserboxh}
\newlength{\teaserboxdp}
\newsavebox{\teaserprobe}

\newcommand{\teasersetcolumns}[1]{%
    \setlength{\teasercolw}{#1}%
    \sbox{\teaserprobe}{\begin{minipage}{\teasercolw}\centering\teaserrefbox{}{}\end{minipage}}%
    \settoheight{\teaserboxh}{\usebox{\teaserprobe}}%
    \settodepth{\teaserboxdp}{\usebox{\teaserprobe}}%
    \addtolength{\teaserboxh}{\teaserboxdp}%
}

\begin{figure*}[t]
    \centering
    \teasersetcolumns{0.23\linewidth}
    \begin{minipage}[t]{0.03\linewidth}
        \vspace{0pt}
        \centering
        \parbox[b][\teaserboxh][c]{\linewidth}{\centering
            \rotatebox{90}{\footnotesize\sffamily Constraints}}\\[4pt]
        \parbox[b][0.879\teasercolw][c]{\linewidth}{\centering
            \rotatebox{90}{\footnotesize\sffamily Generated Motions}}
    \end{minipage}%
    \hfill
    \begin{minipage}[t]{\teasercolw}
        \vspace{0pt}
        \centering
        \teaserrefbox{Heading Direction}{%
            \includegraphics[width=\linewidth]{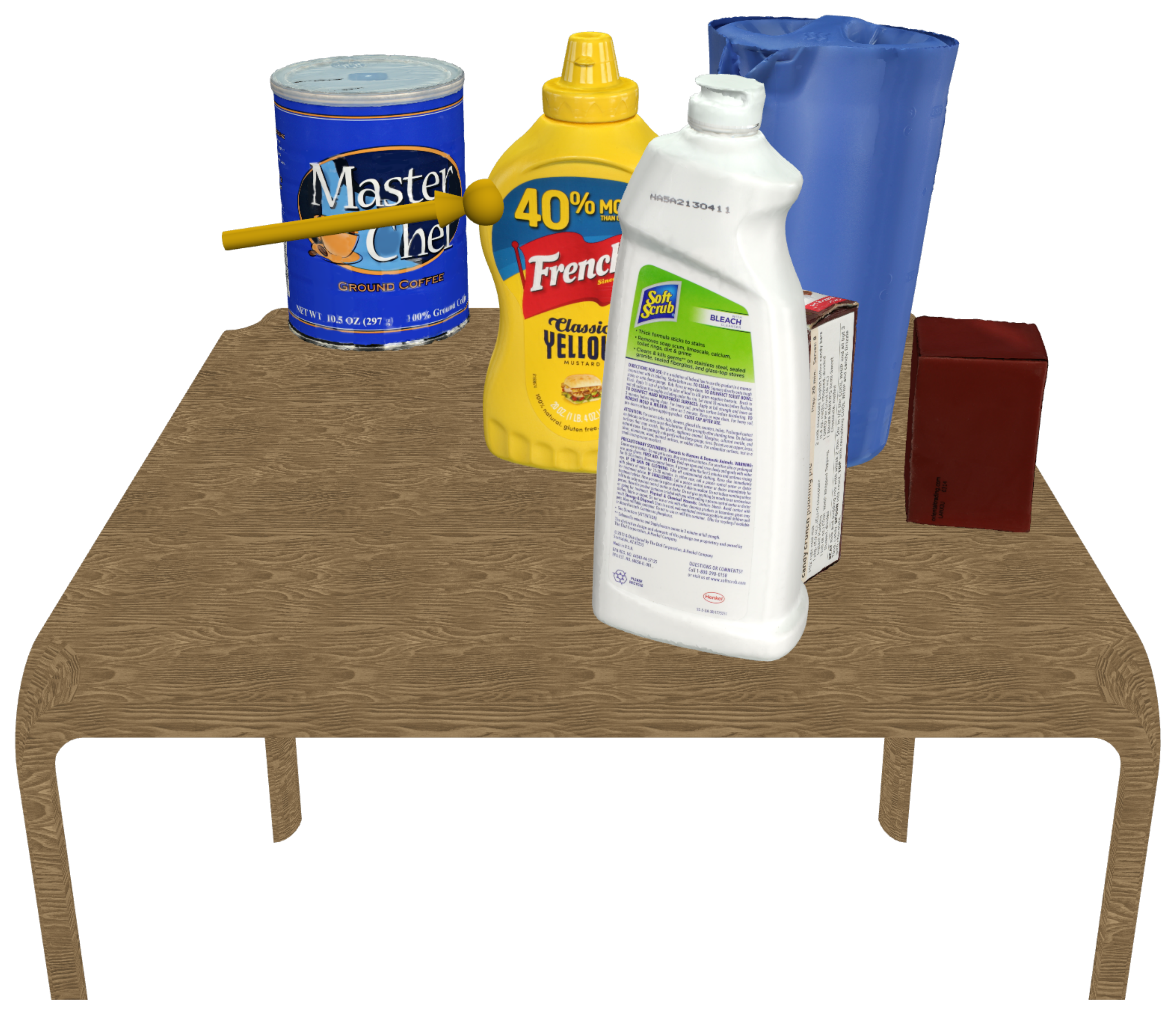}}\\[4pt]
        \teasermotion{\includegraphics[width=\linewidth]{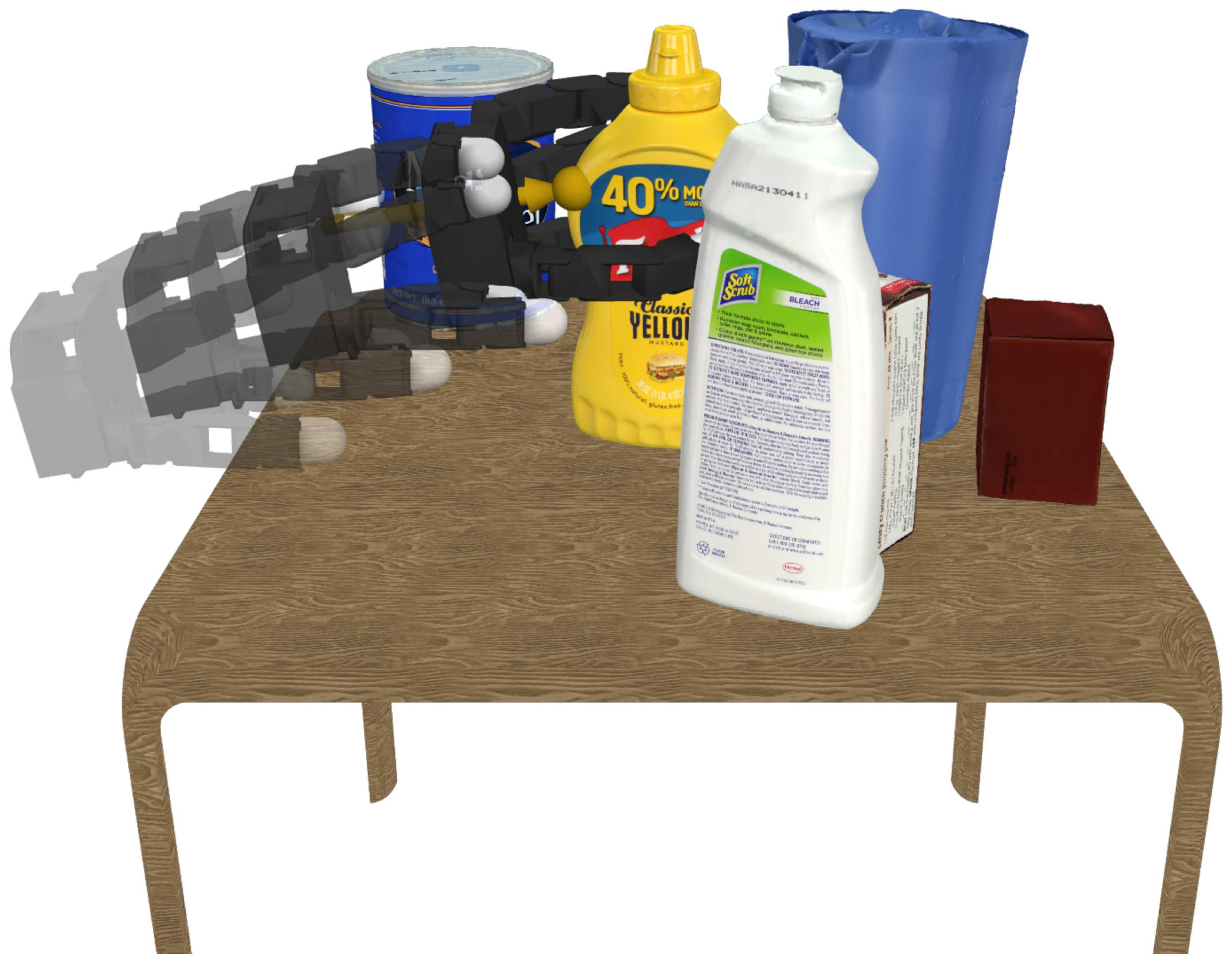}}\\[3pt]
        {\scriptsize\sffamily Grasp from a collision-free direction\par}
    \end{minipage}%
    \hfill
    \begin{minipage}[t]{\teasercolw}
        \vspace{0pt}
        \centering
        \teaserrefbox{Reference Pose}{%
            \includegraphics[height=0.6\linewidth]{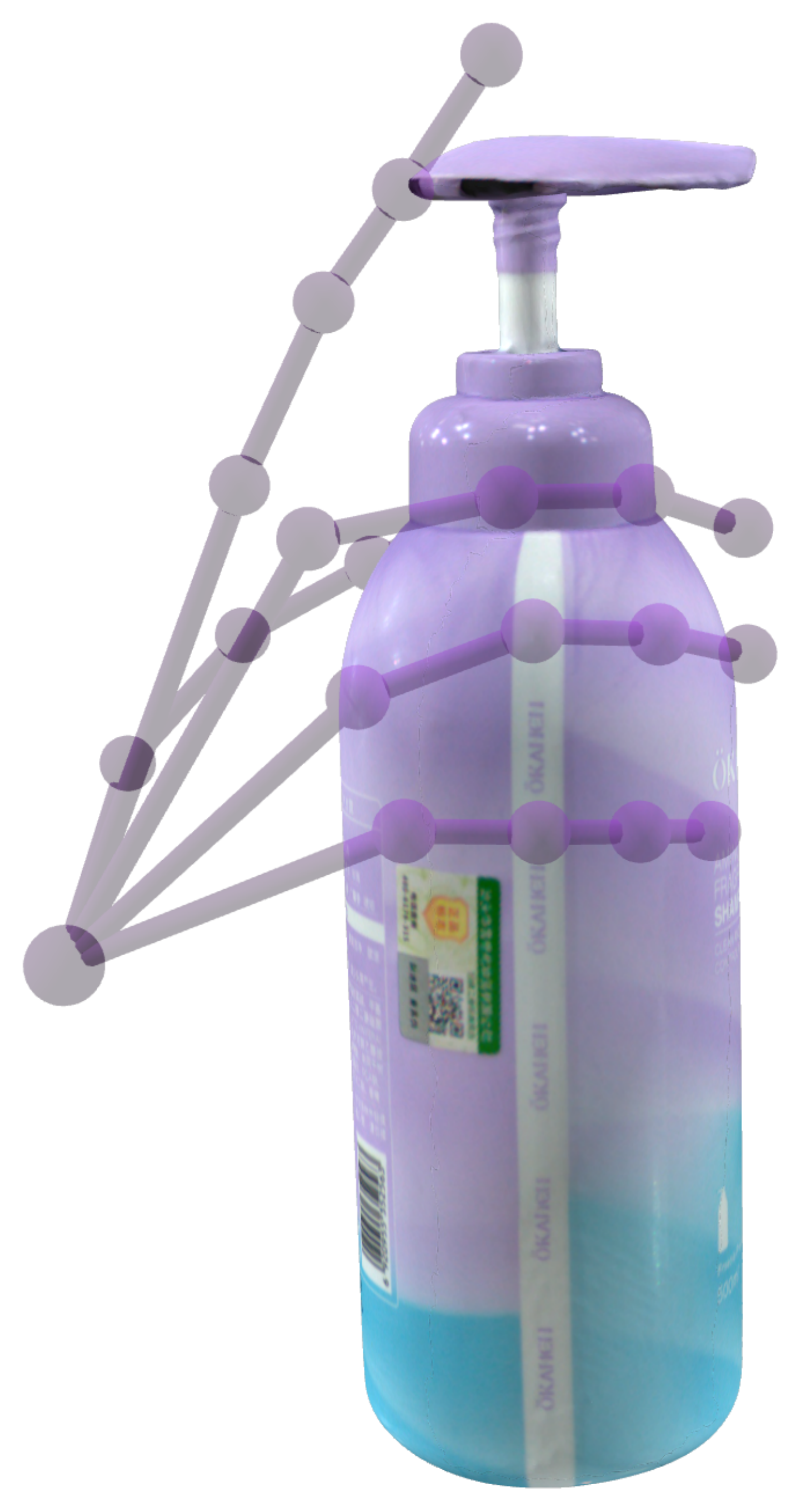}}\\[4pt]
        \teasermotion{\includegraphics[width=\linewidth]{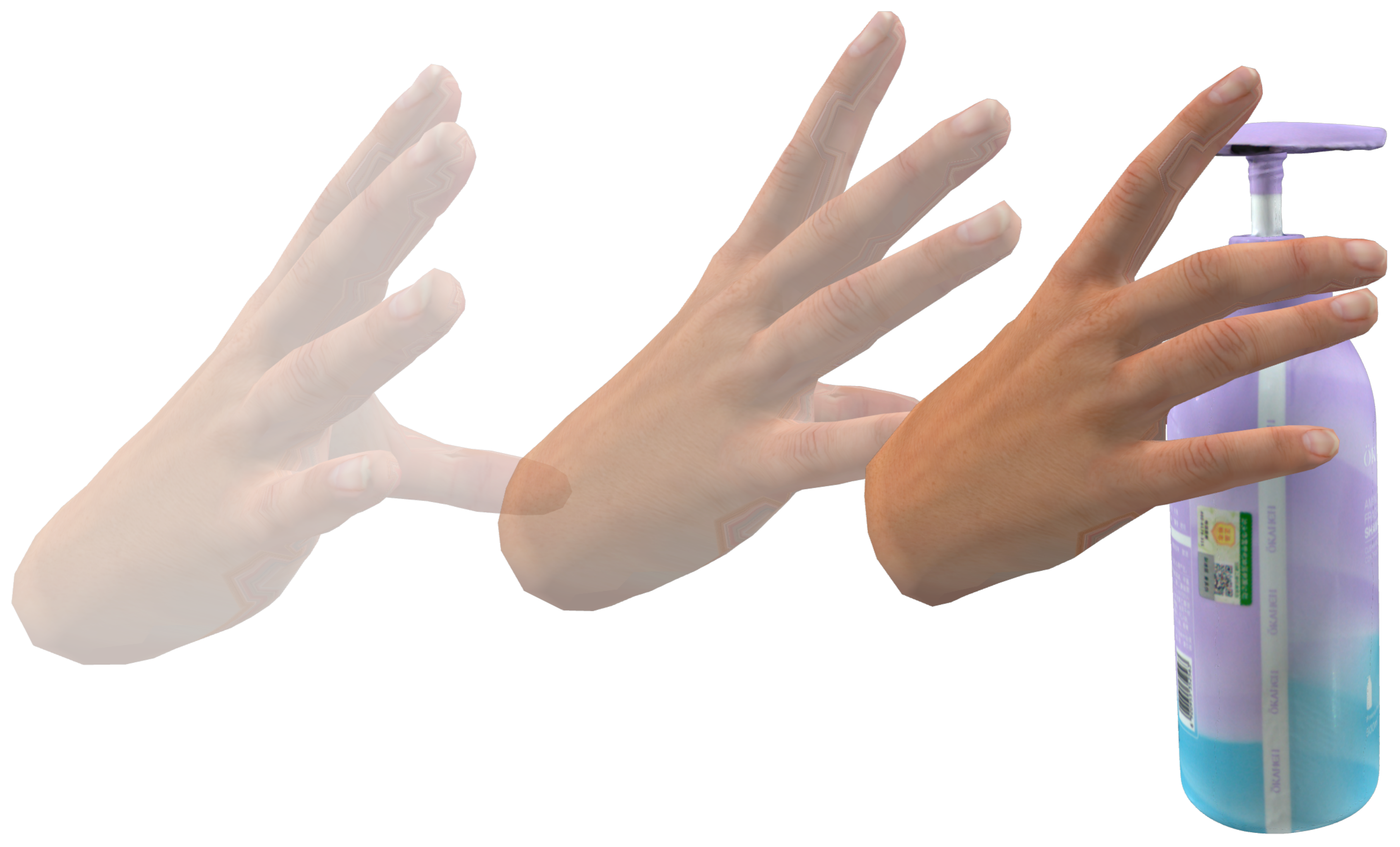}}\\[3pt]
        {\scriptsize\sffamily Generate functional grasping motions\par}
    \end{minipage}
    \hfill
    \begin{minipage}[t]{\teasercolw}
        \vspace{0pt}
        \centering
        \teaserrefbox{Affordance Area}{%
            \includegraphics[width=0.8\linewidth]{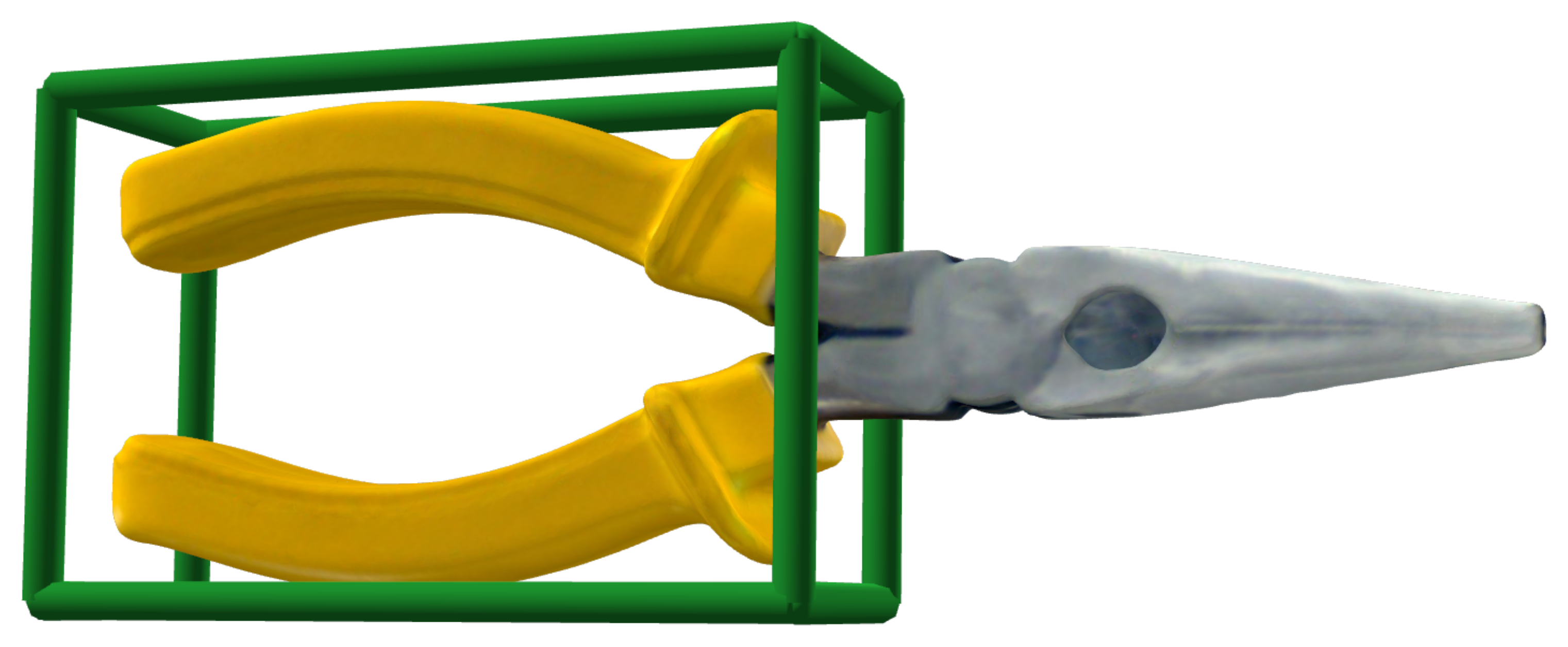}}\\[4pt]
        \teasermotion{\includegraphics[width=\linewidth]{FIG/lossless/aff_seq_tight}}\\[3pt]
        {\scriptsize\sffamily Grasp the desired affordance area\par}
    \end{minipage}%
    \hfill
    \begin{minipage}[t]{\teasercolw}
        \vspace{0pt}
        \centering
        \teaserrefbox{Wrist Trajectory}{%
            \includegraphics[width=\linewidth]{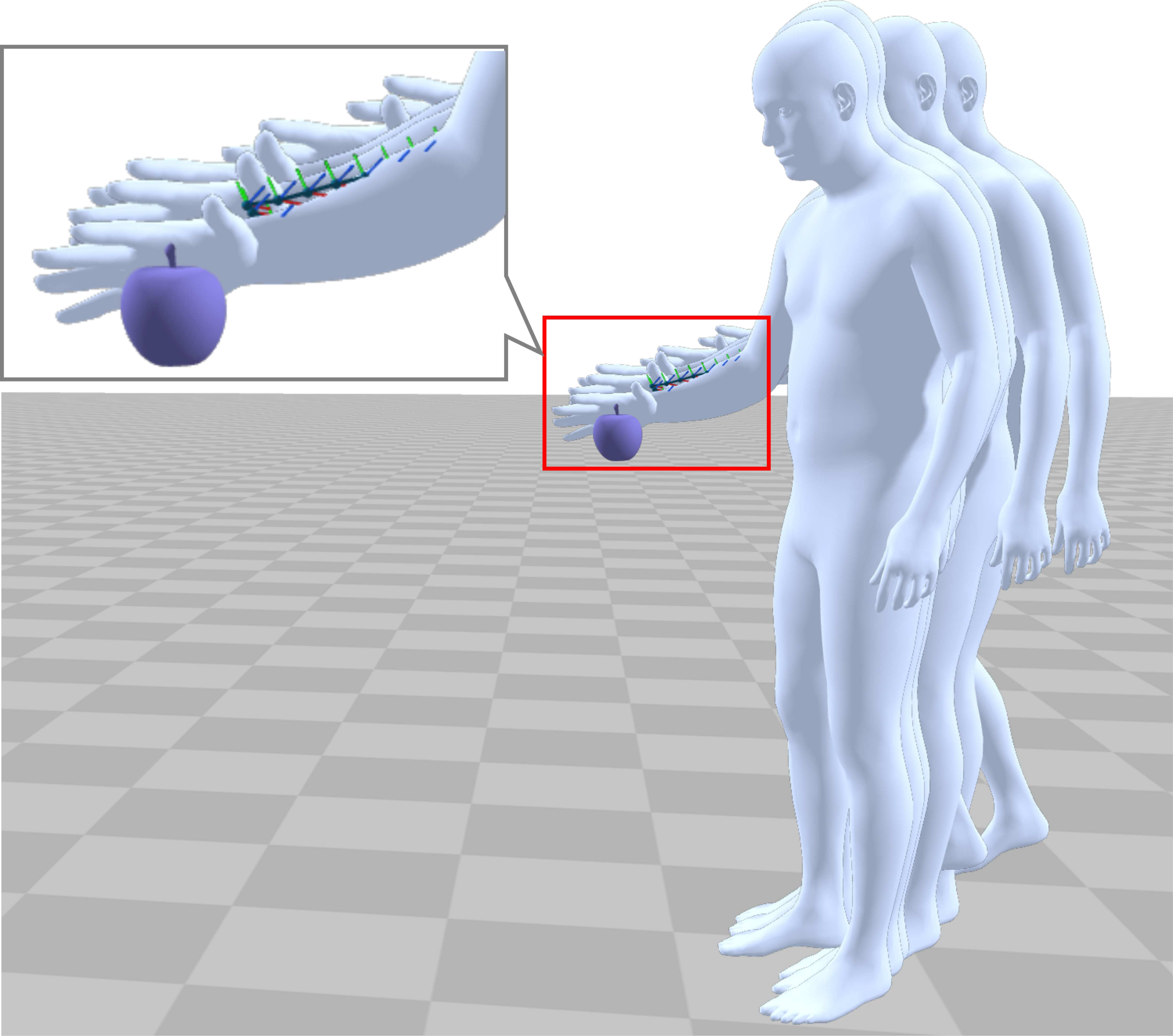}}\\[4pt]
        \teasermotion{\includegraphics[width=\linewidth]{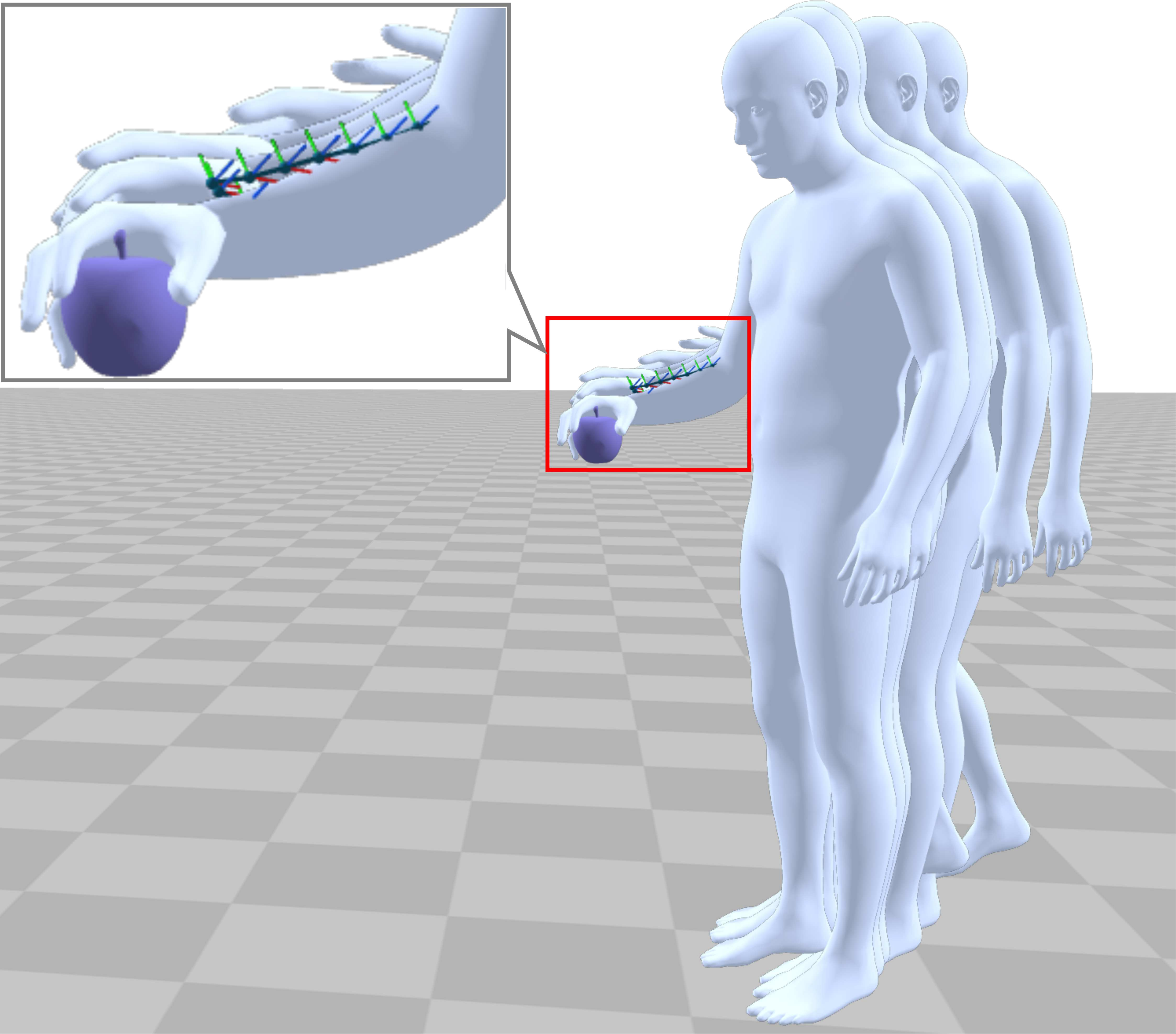}}\\[3pt]
        {\scriptsize\sffamily Fit fingers for whole-body motions\par}
    \end{minipage}%
    \vspace{0.5mm}
    \caption{
    We present \method, a method that can synthesize physics-based grasping motions following various task-driven constraints such as heading directions, affordance, wrist trajectories, and reference poses, serving as a plug-and-play low-level controller for various applications of diverse hand models.
    }
    \vspace{-2mm}
    \label{fig:teaser}
\end{figure*}

%% file: sec/02_related_work.tex
\section{Related Work}

\subsection{Static Dexterous Grasp Synthesis}

Static dexterous grasp synthesis aims to predict a feasible final hand configuration for a given object.
Early methods formulate grasping as an optimization problem~\cite{ye2012synthesis, nguyen1986constructing, miller2004graspit}, usually conditioned on contacts~\cite{brahmbhatt2019contactdb, brahmbhatt2019contactgrasp} or predefined grasping types~\cite{cutkosky1989on, feix2016grasp, corona2020ganhand, chen2025dexonomy}.
Recent works utilize data-driven methods to learn static grasp synthesis from object geometry~\cite{grab, karunratanakul2020grasping}. Some of them predict contacts and then generate or finetune grasps through optimization conditioned on predicted contacts~\cite{li2023contact2grasp, jiang2021graspTTA, liu2023contactgen, yu2026hugs}. 
Instead of synthesizing grasp configurations from scratch, some other works reconstruct human grasping poses from images~\cite{fan2024hold, hasson2019learning, cao2021reconstructing} or retarget human hand grasping poses to different dexterous robot hands~\cite{gavryushin2025maple, mandikal2022dexvip}, leading to fine-grained functional grasping poses.
Overall, these static grasp synthesis methods are useful for generating final configurations. However, they do not model the temporal process of approaching the object, forming contacts, and stabilizing the object under hand dynamics. 
Our work, instead, synthesizes temporal grasping motions and incoperates different dynamic constraints such as heading directions and wrist trajectories during approach. The static grasping poses generated by these methods can also be used as constraints for our temporal grasp motion generation.

\subsection{Dynamic Grasping Motion Synthesis}

Dynamic hand-object interaction synthesis has long been studied in computer vision and graphics~\cite{rijpkema1991computer, pollard2005physically, liu2024geneoh, wang2019learning, diller2024cg, zhang2026unicross}, where grasping motion generation is an important part due to its fundamental role in hand-object interaction~\cite{kalisiak2001grasp, zhao2013robust}.
With the development of hand-object datasets~\cite{chao2021dexycb, fan2023arctic, zhan2024oakink2, liu2024taco, liu2022hoi4d, hoque2025egodex, hampali2020ho3d, kwon2021h2o}, a series of data-driven methods for dexterous interaction motion synthesis have been proposed~\cite{zheng2023cams,cha2024text2hoi, wu2022saga, christen2024diffh2o}. Utilizing the priors extracted from existing datasets, these methods can generate natural motions within the data domain, while their out-of-distribution performance is limited by the scale and diversity of available data. Moreover, some of them require auxiliary signals as mandatory inputs rather than optional conditions during inference, such as wrist trajectories~\cite{zhang2021manipnet} and object motions~\cite{zhang2021manipnet, shimada2023macs, taheri2024grip, li2023object}, which further limits their generalization.

With the development of physics simulation and reinforcement learning, physics-based methods have been applied to reduce this dependence on demonstrations by exploring hand-object interactions directly in simulation, while also improving physical plausibility~\cite{xu2023unidexgrasp, wang2025learning, christen2022dgrasp}.
Some methods use a static reference grasp pose to generate dynamic motions~\cite{christen2022dgrasp, Wan_2023_ICCV, zhang2024artigrasp}.
Our prior work, GraspXL~\cite{zhang2024graspxl}, takes this direction further by generating dynamic grasping motions without relying on hand-object interaction data, leading to extraordianry generalization to 500k+ unseen objects and diverse hand morphologies.
In this paper, we preserve the large-scale generalization ability of GraspXL and specifically focus on controllability for task-driven, constraint-conditioned grasp motion generation.
The resulting policy can serve as a plug-and-play low-level generator that produces physically stable grasping motions tailored to heterogeneous constraints from different tasks.

\subsection{Controllable Grasping Motion Synthesis}

Constraint-conditioned motion generation has been shown effective for task-oriented full-body human motion synthesis conditioned on sparse or partial signals, such as key frames, body trajectories, key point positions, and semantic intents~\cite{tessler2024maskedmimic, xie2024omnicontrol, shafir2024human, karunratanakul2023guided}. Achieving such controllability to dexterous grasping, however, is more challenging. Unlike full-body motion generation, dexterous grasping requires precise alignment between the hand and the object movements, stable contact formation, and physical plausibility such as penetration avoidance, while task-driven constraints demand even more fine-grained control to precisely meet the task requirements.

Different lines of work can provide useful task-level constraints for dexterous grasping. For example, affordance can be obtained from visual affordance prediction or semantic part segmentation~\cite{do2018affordancenet, qian2024affordancellm, nguyen2023languageconditioned, nagarajan2019grounded, li2025learning}. Functional static grasp configurations can be generated, reconstructed, or retargeted from images or human demonstrations~\cite{chen2025web2grasp, hampali2020honnotate, antotsiou2018task}. Wrist or hand trajectories can be provided by whole-body motion generation methods~\cite{xu2025intermimic, pan2025tokenhsi}. However, due to the diversity of constraint types and the control complexity of dexterous grasping, existing methods are usually designed for limited and isolated control signals, such as affordance-conditioned grasping~\cite{mandikal2020graff, ye2023affordance, zhao2025afforddex} and reference-pose-conditioned functional grasping~\cite{huang2025fungrasp, agarwal2023dexterous}. In this work, we focus on more universal and flexible grasping motion synthesis conditioned on heterogeneous constraint and their combinations, enabling physically stable grasping motions synthesis under different task requirements.

%% file: sec/03_method.tex
\section{Method}
\label{sec:method}

We model the controllable constraint-conditioned dexterous grasp motion generation problem as a Markov decision process (MDP) solved with PPO~\cite{schulman2017proximal} in a physics simulator. Specifically, given the object $\mathcal{O}$, the dexterous hand $\mathcal{H}$, and the task-driven constraints $\mathcal{T}$, the policy must synthesize a physically stable grasping motion for both the hand and object following the active constraints.
As an extension of GraspXL~\cite{zhang2024graspxl}, this section specifically focuses on the components that differ from GraspXL. Components retained from GraspXL are explicitly noted and briefly described.
We first introduce the hand modeling in Sec.~\ref{sec:hand}. Then we formulate the hierarchical constraints in Sec.~\ref{sec:constraint}, describe the observation space in Sec.~\ref{sec:obs}, the control space in Sec.~\ref{sec:control}, and the reward function in Sec.~\ref{sec:reward}.

\subsection{Hand Modeling}
\label{sec:hand}

\input{figures/heading}

We model a dexterous hand $\mathcal{H}$ as an articulated structure of $L$ links and $N$ joints, where $N=F+6$, with $F$ finger joints and $6$ virtual wrist joints emulating the wrist's six degrees of freedom. 
Following GraspXL, we attach a local grasping coordinate system to the hand, as illustrated in Fig.~\ref{fig:heading}. Specifically, with an open hand, the cross-finger axis $\textbf{w}$ (red) points from the thumb fingertip toward the mean position of the non-thumb fingertips, and the heading axis $\textbf{v}$ (green) points forward from the midpoint of these two endpoints and is orthogonal to $\textbf{w}$. 

We further define a hand center $\textbf{p}$ to represent the grasping center of the hand. 
Unlike GraspXL, which fixes the hand center at a point in the wrist frame determined by the initial hand pose, we adopt a \emph{dynamic hand center} defined as the real-time midpoint between the mean non-thumb fingertips and the thumb fingertip. 
As illustrated in Fig.~\ref{fig:dynamic_hand_center}, this dynamic definition adapts the grasping center with different depths as the fingers open and close, guiding the fingertips to adaptively wrap the object of different sizes. Moreover, for the MANO~\cite{MANO:SIGGRAPHASIA:2017} hand in particular, we build a hand model with $20$ anatomical finger joints ($F=20$, four per finger, each equipped with anatomical joint limits and rotation axes). This yields greater anatomical plausibility than GraspXL, which models $45$ finger joints directly from the MANO template without anatomical meaning and thus require an additional anatomical loss~\cite{yang2021cpf} during training as a soft constraint.

\input{figures/dynamic_hand_center}

\subsection{Hierarchical Constraint Formulation}
\label{sec:constraint}

Different downstream tasks impose different requirements on how an object should be grasped, at different levels of details.
A pick-and-place task may only require the hand to approach the object from a collision-free direction, imposing a coarse heading constraint.
A tool-use task might require a functional grasp that contacts only the handle, imposing affordance and pose constraints.
A task coupled with whole-body motion might require the wrist to follow a planned path specified by the whole-body motion, imposing a fine-grained wrist trajectory constraint.
Such requirements may be specified directly by a user or produced by off-the-shelf modules, e.g., an affordance model learned from human-object interaction data, a human-to-robot pose-retargeting pipeline, or a whole-body motion generator.

To serve these diverse task requirements, we organize the task-driven constraints into four levels, namely heading constraints, affordance constraints, pose constraints, and trajectory constraints, ranging from coarse cues such as the approach direction and contact regions to fine-grained specifications such as the wrist trajectory and finger poses. 
Since these constraints describe how the hand should interact with the object, they are all defined in the object frame.

Heading constraints define the desired approach behavior of the hand to the object, which is represented as a heading direction $\bar{\textbf{v}}$ and a heading position $\bar{\textbf{p}}$ in the object frame as shown in Fig.~\ref{fig:heading}. Ideally, the wrist-frame heading axis $\textbf{v}$ should be parallel to $\bar{\textbf{v}}$ to approach the object from the desired direction, while the hand center $\textbf{p}$ should be aligned with $\bar{\textbf{p}}$ to wrap the object around the desired position.

Reference pose constraints describe desired instantaneous properties of the hand or wrist, which is represented as a target 6D wrist pose $\bar{\textbf{T}}$ in the object frame and finger joint angles $\bar{\textbf{q}}$, together with the finger link positions $\bar{\textbf{x}}$ in the object frame. 

Wrist trajectory constraints describe the desired temporal behavior of the wrist as a sequence of poses $\bar{\textbf{S}}=\{\bar{\textbf{T}}_{1},\ldots,\bar{\textbf{T}}_{K}\}$.

Affordance constraints specify where the hand should and should not interact with the object through an affordance partition, which is represented as a desired grasping region $\mathcal{O}^{+}$ and a non-grasping region $\mathcal{O}^{-}$, with $\mathcal{O}=\mathcal{O}^{+}\cup\mathcal{O}^{-}$ is the object surface point cloud.

\input{figures/constraint_priority}

As these constraints specify grasping at different levels, they may overlap on the same control intention. We therefore introduce a priority hierarchy as shown in Fig.~\ref{fig:constraint_priority}: when constraints overlap, the one that more explicitly or detailedly specifies the intended motion takes precedence.
Specifically, the wrist trajectory explicitly specifies the temporal evolution of the wrist pose so it overrides the heading constraint as well as the wrist component of the reference pose, while the finger component of the reference pose remains active. 
The reference pose, in turn, provides a more specific wrist target and contact regions than the heading and affordance constraints and thus overrides them. 
The affordance overrides the heading only when the specified direction points closer to the non-grasping region $\mathcal{O}^{-}$ than to the grasping region $\mathcal{O}^{+}$, in which case it redirects the heading toward $\mathcal{O}^{+}$. 
Trajectory and affordance constraints do not override one another. Instead, they should be mutually aligned so that the final wrist position of the trajectory constraint is closer to the grasping region $\mathcal{O}^{+}$ than the non-grasping region $\mathcal{O}^{-}$.
The heading forms the lowest-priority default and is the only constraint that is always active. When no task-specific heading is provided, we sample a random heading direction $\bar{\textbf{v}}$ and set the heading position $\bar{\textbf{p}}$ to the object center.

\input{figures/framework}

\subsection{Observation Space}
\label{sec:obs}
As illustrated in Fig.~\ref{fig:framework}, at each time step $t$, the observation consists of states $\textbf{s}_t$ and constraints $\mathcal{T}_t$. The state representation follows GraspXL and captures the task-relevant information of the current hand and object states, while the constraint representation encodes the active task-driven constraints introduced in Sec.~\ref{sec:constraint}.

\subsubsection{State representation}
The hand state summarizes the proprioceptive and contact status of the hand $\mathcal{H}$. Concretely, it comprises the current finger joint angles $\textbf{q}_t\in\mathbb{R}^{F}$ and virtual wrist joint angles $\textbf{u}_t\in\mathbb{R}^{6}$ together with their previous target angles $\textbf{q}^{\text{tgt}}_{t-1}\in\mathbb{R}^{F}$ and $\textbf{u}^{\text{tgt}}_{t-1}\in\mathbb{R}^{6}$, and, for each link $i\in\{1,\ldots,L\}$, a binary contact flag $b^i_t\in\{0,1\}$ and the corresponding contact force $\textbf{f}^i_t\in\mathbb{R}^{3}$.
Similar as in GraspXL, the object is represented in an interaction-driven manner with the nearest-point distance vectors $\{\textbf{d}^i_t\}_{i=1}^{L}$ from the hand links to the object surface points $\mathcal{O}$, expressed in the wrist frame. 

\subsubsection{Constraint representation}
The constraint representation encodes the four constraint levels of Sec.~\ref{sec:constraint} as features relative to the current hand state. Since constraints can be separately activated, each level (except for the heading level which is always active) is gated by a binary activity mask $m\in\{0,1\}$ that zeros out its features when the corresponding constraint is absent, so a missing constraint contributes a zero feature rather than a spurious target. 

The heading level constraint provides the hand center $\textbf{p}_t$, the heading position $\bar{\textbf{p}}$, and their residual $\bar{\textbf{p}}-\textbf{p}_t$, together with the heading direction $\bar{\textbf{v}}$, all expressed in the wrist frame. As $\bar{\textbf{p}}$ and $\bar{\textbf{v}}$ are imposed in the object frame (which direction should the hand approach the object from), we also involve the object orientation $\textbf{R}_t$ in the wrist frame in the observation space so that the policy can reason about how the wrist should move to achieve the heading constraint.

The affordance level constraint captures the shape feature of the non-grasping region $\mathcal{O}^{-}$ by the nearest-point distance vectors $\{\textbf{d}^{i,-}_t\}_{i=1}^{L}$ from the hand links to the non-grasping region $\mathcal{O}^{-}$, expressed in the wrist frame, and the net contact force of each link $\textbf{f}^{i,-}_t$ acting on it, so the policy can reason about the geometry to avoid.

The pose level constraint captures the desired instantaneous properties of the hand or wrist by the finger-joint residual position $\bar{\textbf{q}}-\textbf{q}_t$ and the finger-link position residuals $\bar{\textbf{x}}-\textbf{x}_t$ in the wrist frame, together with the residual virtual wrist joint angles $\bar{\textbf{u}}-\textbf{u}_t$ calculated from the wrist pose difference to the target $\bar{\textbf{T}}$. The finger and wrist parts are masked separately so that they can be activated or overridden independently.

The trajectory level constraint captures the desired temporal behavior of the wrist by a sequence of residual virtual wrist joint angles $\{\bar{\textbf{u}}^{(k)}-\textbf{u}_t\}_{k=1}^{K}$, each calculated from the wrist pose difference to the $k$-th target frame $\bar{\textbf{T}}_{k}$ of the wrist trajectory $\bar{\textbf{S}}$.

\subsection{Control Space}
\label{sec:control}

At each step, the policy outputs a residual action $\mathbf{a}_t=[\mathbf{a}^{f}_t,\mathbf{a}^{w}_t]$, split into a finger part $\mathbf{a}^{f}_t\in\mathbb{R}^{F}$ and a virtual-wrist part $\mathbf{a}^{w}_t\in\mathbb{R}^{6}$, which are decoded into finger and wrist joint targets respectively.

\subsubsection{Finger control}
For the fingers, the residual action is scaled and added to the current finger target angles,
\begin{equation}
    \mathbf{q}^{\text{tgt}}_{t+1} = \text{clamp}\big(\mathbf{q}^{\text{tgt}}_{t} + \boldsymbol{\alpha}^{f}\cdot\mathbf{a}^{f}_t,\ \mathbf{q}_{\min},\mathbf{q}_{\max}\big),
\end{equation}
where $\boldsymbol{\alpha}^{f}$ is the finger-joint action scale and $\text{clamp}(\cdot)$ enforces the anatomical joint limits $[\mathbf{q}_{\min},\mathbf{q}_{\max}]$. This incremental parameterization yields temporally coherent and smooth motions.

\subsubsection{Feed-forward wrist guidance}
GraspXL controls the wrist with a simple hand guidance that biases the wrist joints by the difference between the heading-derived and current wrist poses.
To serve the heterogeneous constraints, we generalize this into a \emph{feed-forward wrist guidance} in which the wrist residual action is applied on top of a real-time guidance target instead of a fixed heading-derived target pose,
\begin{equation}
    \mathbf{u}^{\text{tgt}}_{t+1} = \mathbf{u}^{g}_t + \boldsymbol{\alpha}^{w}\cdot\mathbf{a}^{w}_t,
\end{equation}
where $\mathbf{u}^{g}_t$ are the virtual wrist joint angles converted from a guidance wrist pose $\bar{\textbf{T}}^{g}_t$. Following the constraint hierarchy (Sec.~\ref{sec:constraint}), $\bar{\textbf{T}}^{g}_t$ is resolved from the highest active wrist constraint: with only a heading constraint it is the heading-derived target wrist pose, which recovers the GraspXL behavior; with an active reference pose it is the target wrist pose $\bar{\mathbf{T}}$; and with an active wrist trajectory it is the waypoint after the one closest to the current wrist position, i.e., $\bar{\textbf{T}}^{g}_t=\bar{\textbf{T}}_{k^{*}+1}$ with $k^{*}=\arg\min_{k}\|\mathbf{t}^{w}_t-\bar{\mathbf{t}}^{w}_{k}\|_2$, where $\bar{\mathbf{t}}^{w}_{k}$ is the $k$-th waypoint in the trajectory and $\mathbf{t}^{w}_t$ is the current wrist position. This feed-forward design encourages the policy to advance forward along the trajectory with continuous motion.

The resulting target joint positions $\mathbf{q}^{\text{tgt}}_{t+1}$ and $\mathbf{u}^{\text{tgt}}_{t+1}$ are then converted to joint torques by a low-level PD controller and applied to the joints to drive the desired motion in simulation.

\subsection{Reward Function}
\label{sec:reward}

We design a morphologically-agnostic reward function that encourages the policy to form stable grasps while satisfying the constraints. The overall reward structure is similar as GraspXL with $r_t = r_t^{\text{grasp}} + r_t^{\text{track}}$, while the specific terms and formulation are adapted to the heterogeneous constraints.
All reward weights are provided in the supplementary material.

\subsubsection{Grasp reward}

The grasp reward encourages the fingers to approach the object surface and maintain consistent contacts with the object, while avoiding extreme motions:

\begin{equation}
    r_t^{\text{grasp}} = r_t^{\text{dis}} + r_t^{\text{con}} + r_t^{\text{reg}}
\end{equation}

The distance term $r_t^{\text{dis}} = -w_{\text{dis}}\sum_{i=1}^{L}\|\textbf{d}^i_t\|$ pulls the hand links toward the object surface.
The contact term $r_t^{\text{con}} = \sum_{i=1}^{L} b^i_t\,(w_c+w_f\min(\|\textbf{f}^i_t\|, f_{\max}))$ rewards stable contacts while clamping the contact force at $f_{\max}$ to prevent excessive forces.
The regularization term $r_t^{\text{reg}} = -w_{o}\|\textbf{m}^o_t-\textbf{m}^o_0\|^2 - w_{R}\,\mathcal{A}(\textbf{R}^o_t,\textbf{R}^o_0) - w_{u}\|\dot{\textbf{u}}_t\|^2 - w_{q}\|\textbf{q}_t-\textbf{q}_0\|^2 - w_{\text{drop}}\,\mathbb{I}_\text{drop}$ penalizes object displacement in both position $\textbf{m}^o_t$ and orientation $\textbf{R}^o_t$ from their initial values $\textbf{m}^o_0$ and $\textbf{R}^o_0$, wrist velocity $\dot{\textbf{u}}_t$ (the velocity of the six virtual wrist joints), deviation from the initial finger pose $\textbf{q}_0$, and object drop, where $\mathcal{A}(\cdot,\cdot)$ is the squared angular difference between two rotations and $\mathbb{I}_\text{drop}$ indicates whether the object has fallen.

Notably, with the anatomically based hand model (see Sec.~\ref{sec:hand}), we eliminate the additional anatomical penalty term required for MANO in GraspXL, yielding a more consistent reward formulation across different hands. 

\subsubsection{Constraint tracking reward}

The constraint tracking reward drives the policy to satisfy the active task-driven constraints, aggregating four terms that cover the heading, pose, trajectory, and affordance levels:
\begin{equation}
    r_t^{\text{track}} = r_t^{\text{head}} + r_t^{\text{pose}} + r_t^{\text{traj}} + r_t^{\text{aff}}.
\end{equation}
Following the constraint representation, each term is gated by the activity mask of its level, so an inactive constraint contributes zero reward.

The heading term encourages the heading axis $\textbf{v}_t$ and the hand center $\textbf{p}_t$ to be aligned with the desired values $\bar{\textbf{v}}$ and $\bar{\textbf{p}}$:
\begin{equation}
    r_t^{\text{head}} = w_{\text{dir}}\textbf{v}_t^\top\bar{\textbf{v}} - w_{\text{pos}}\left\|\textbf{p}_t-\bar{\textbf{p}}\right\|_2,
\end{equation}
where $\textbf{v}_t$ is the current heading axis and $\textbf{p}_t$ is the hand center, both expressed in the wrist frame, and $\bar{\textbf{v}}$ and $\bar{\textbf{p}}$ are the target heading direction and position.
 (assigned in the object frame and converted to the wrist frame). 

The pose term encourages the current hand configuration to match the reference in the joint, link, and wrist spaces:
\begin{equation}
    \begin{aligned}
    r_t^{\text{pose}}
    ={}&-g(d_t)\left(
    w_q\|\textbf{q}_t-\bar{\textbf{q}}\|^2
    +w_x\|\textbf{x}_t-\bar{\textbf{x}}\|^2
    \right)\\
    &-w_p\|\textbf{t}^w_t-\bar{\textbf{t}}^w\|^2
    -w_R\mathcal{A}(\textbf{R}^w_t,\bar{\textbf{R}}^w),
    \end{aligned}
    \end{equation}
  
where $g(d_t)$ linearly increases from $0$ to $1$ as the distance $d_t$ between the hand and the object decreases from $0.1$m to $0.03$m.
$\textbf{q}_t$ is the finger joint angles, $\textbf{x}_t$ is the wrist-frame finger-link positions, and $(\textbf{t}^{w}_t,\textbf{R}^{w}_t)$ the root-frame wrist pose, tracking the reference $(\bar{\textbf{q}},\bar{\textbf{x}},\bar{\textbf{t}}^{w},\bar{\textbf{R}}^{w})$ which is assigned in the object frame and converted to the wrist and root frames, respectively.

The trajectory term encourages the wrist to follow the planned approach path instead of only matching the final pose:

\begin{equation}
    r_t^{\text{traj}}
    = - w_{\text{traj-p}}\|\textbf{t}^{w}_t-\bar{\textbf{t}}^{w}_{k}\|_2^2
    - w_{\text{traj-q}}\,\mathcal{A}(\textbf{R}^{w}_t,\bar{\textbf{R}}^{w}_{k}),
\end{equation}
where $(\bar{\textbf{t}}^{w}_{k},\bar{\textbf{R}}^{w}_{k})$ is the active waypoint of $\bar{\textbf{T}}^{g}_t$ selected as described in Sec.~\ref{sec:control}. 

The affordance term penalizes both proximity to and forces on the non-grasping region $\mathcal{O}^{-}$:
\begin{equation}
    r_t^{\text{aff}}
    =
    -
    w_{d}\operatorname{softplus}\!\Big(\tfrac{m-d^{-}_t}{\tau}\Big)
    -
    w_{f}\|\textbf{f}^{-}_t\|,
\end{equation}
where $d^{-}_t$ is the distance from the fingers to the nearest surface point of $\mathcal{O}^{-}$, $m$ is the distance margin, and $\tau$ controls the sharpness of the near-surface barrier. $\textbf{f}^{-}_t$ is the per-link contact force on $\mathcal{O}^{-}$. We set $m=0.01$ m and $\tau=0.002$ m, which means the policy is encouraged to keep the fingers at least 0.01 m away from the non-grasping region.

%% file: figures/heading.tex
\begin{figure}[t]
    \centering
    \includegraphics[width=0.75\linewidth]{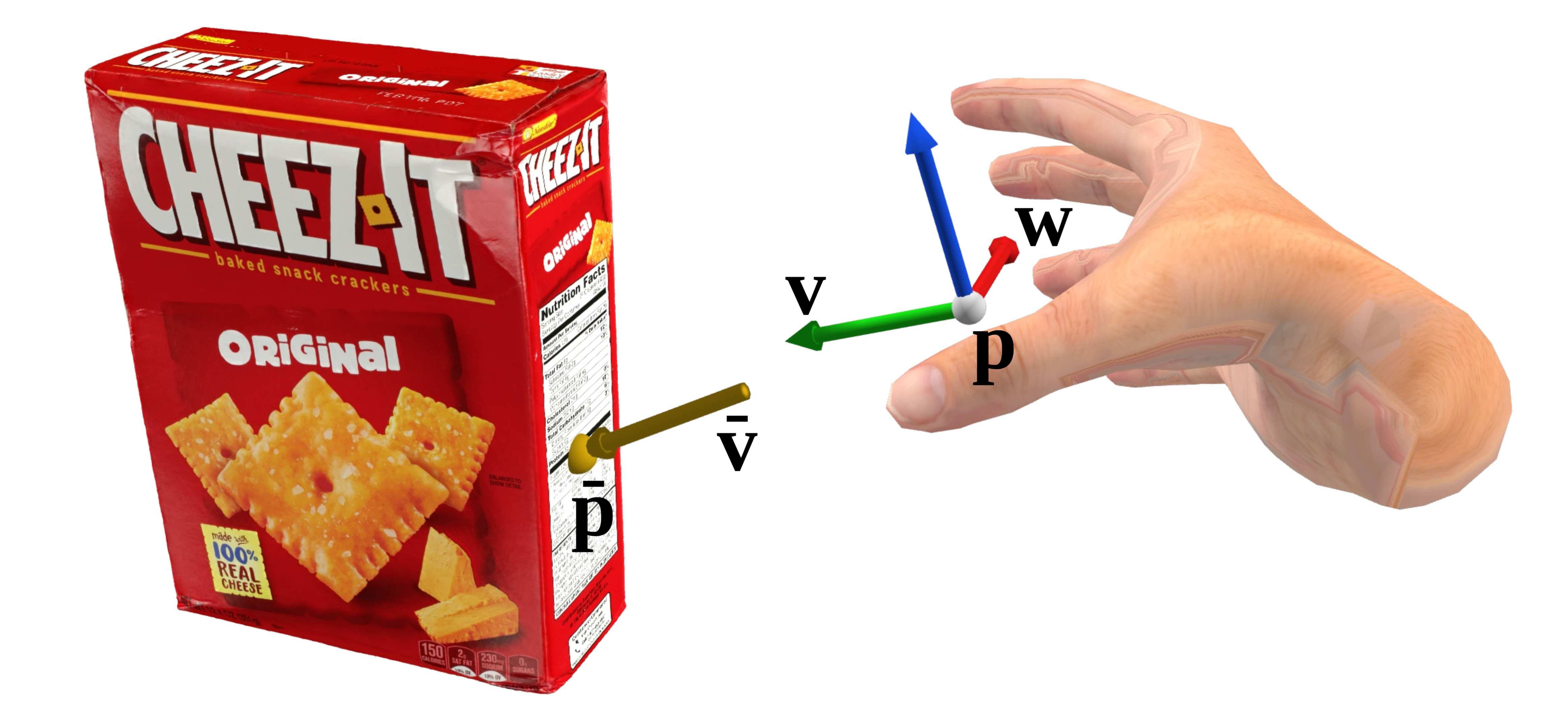}
    \caption{
    \textbf{Grasping Coordinates and Heading Constraints.}
    The cross-finger axis $\textbf{w}$ (red) and heading axis $\textbf{v}$ (green) form the local grasping coordinate frame. $\textbf{p}$ is the hand center.
    The heading constraint specifies a desired heading direction $\bar{\textbf{v}}$ and position $\bar{\textbf{p}}$ in the object frame.
    }
    \label{fig:heading}
\end{figure}

%% file: figures/dynamic_hand_center.tex
\begin{figure}[t]
    \centering
    \includegraphics[height=0.45\linewidth]{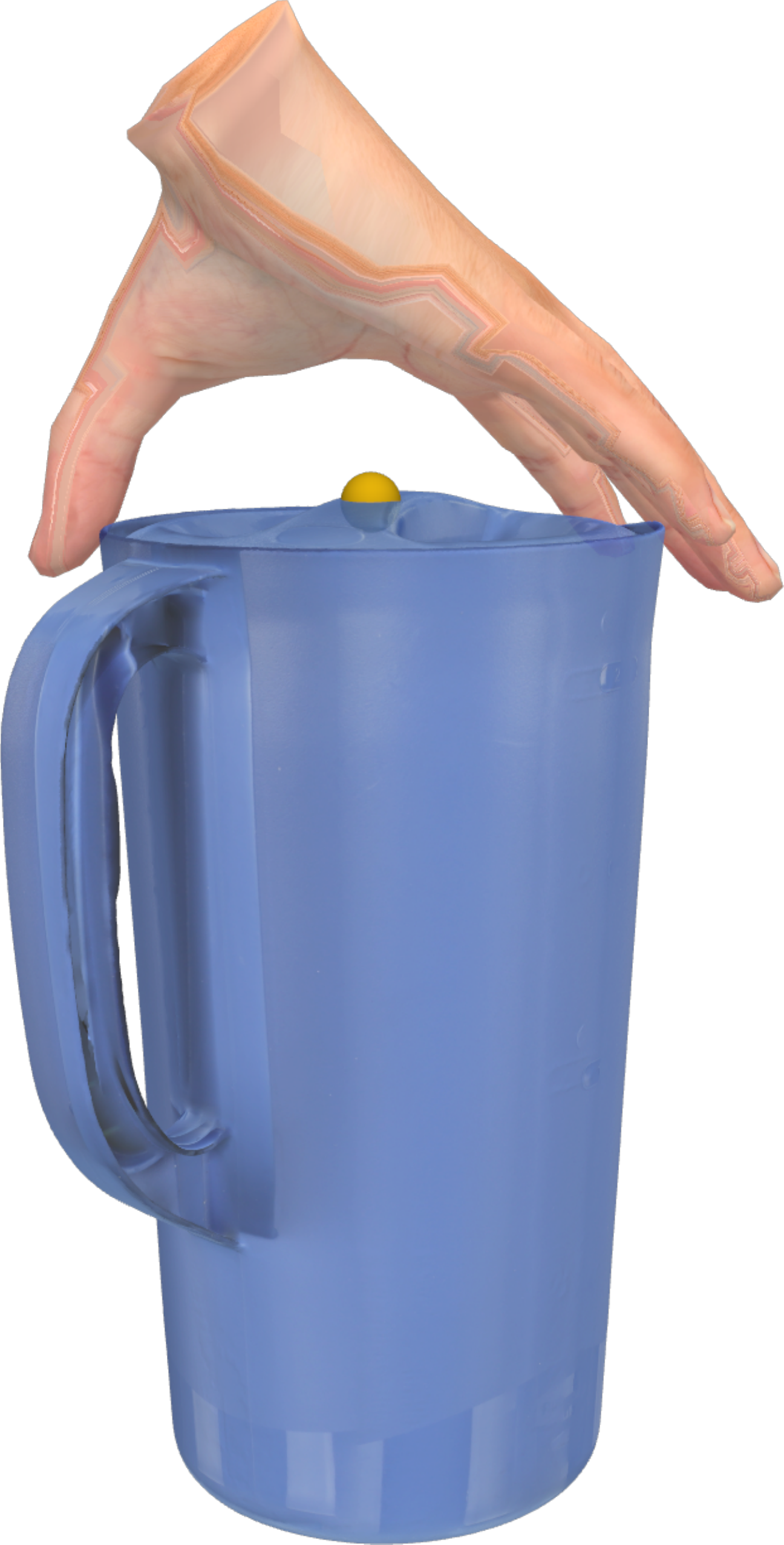}%
    \hspace{0.3\linewidth}%
    \includegraphics[height=0.27\linewidth]{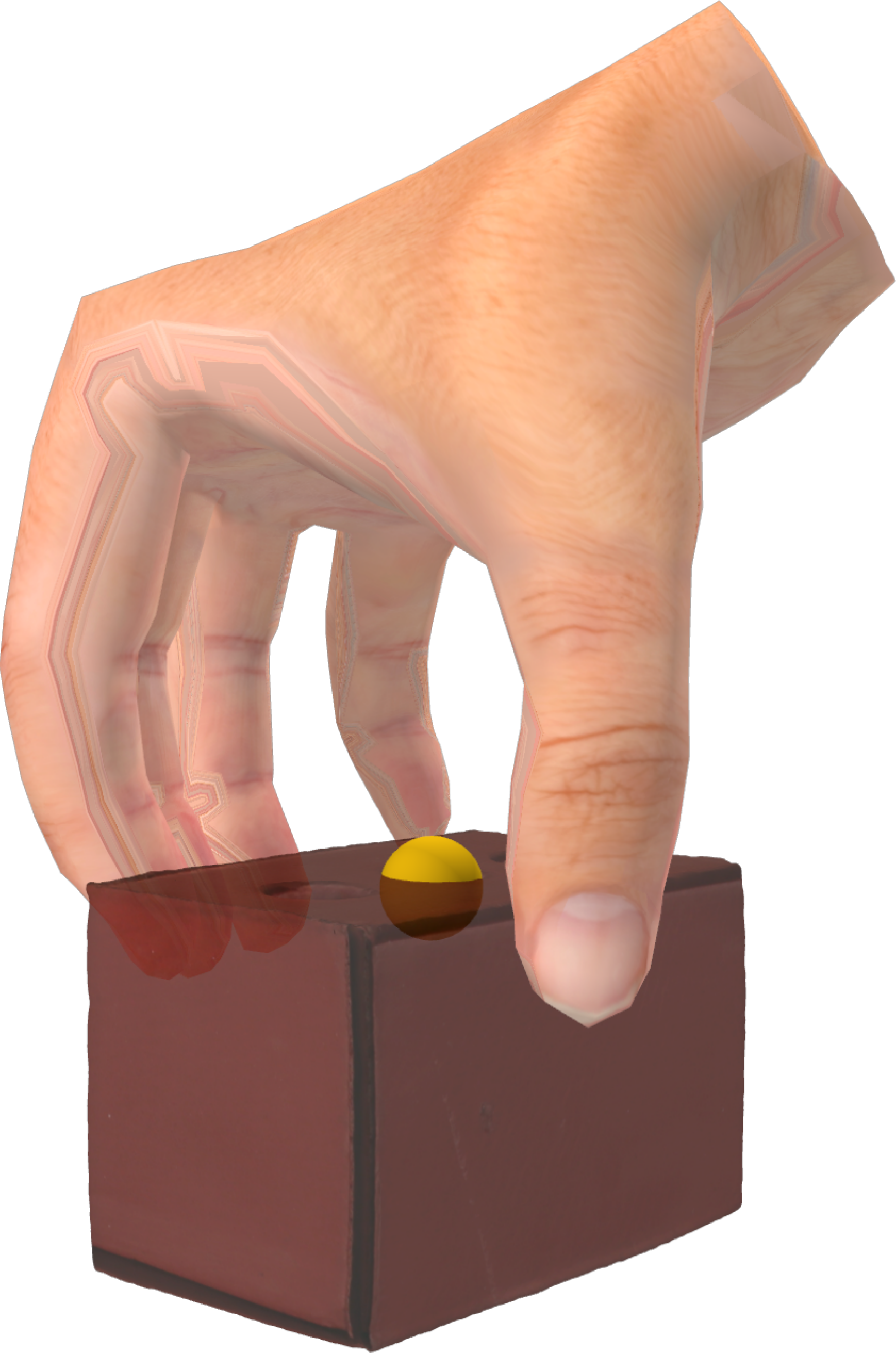}
    \caption{
    \textbf{Dynamic Hand Center.}
    A dynamic hand center (the yellow dot) can adaptively encourage the fingertips to wrap the object of different sizes as the fingers open or close.
    }
    \vspace{-2mm}
    \label{fig:dynamic_hand_center}
\end{figure}

%% file: figures/constraint_priority.tex
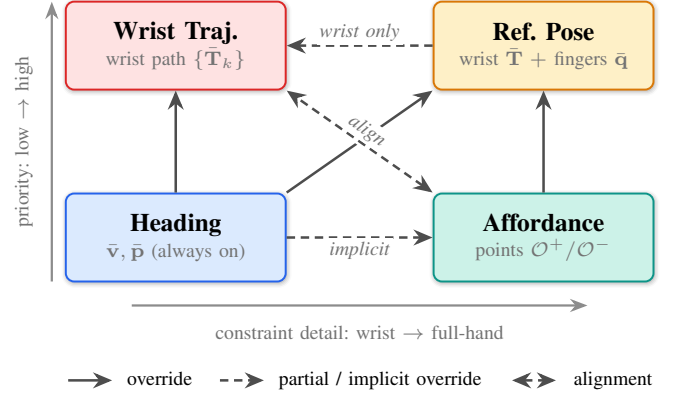
\begin{figure}[t]
    \centering
    \definecolor{cHeadBg}{RGB}{219,234,254}\definecolor{cHeadFg}{RGB}{37,99,235}
    \definecolor{cAffBg}{RGB}{204,242,232}\definecolor{cAffFg}{RGB}{13,148,136}
    \definecolor{cPoseBg}{RGB}{254,240,199}\definecolor{cPoseFg}{RGB}{217,119,6}
    \definecolor{cTrajBg}{RGB}{254,226,226}\definecolor{cTrajFg}{RGB}{220,38,38}
    \tikzset{
        cbox/.style={rounded corners=3pt, draw, line width=0.7pt,
            minimum width=2.75cm, minimum height=1.1cm, align=center,
            font=\small, inner sep=2pt,
            blur shadow={shadow blur steps=4, shadow xshift=0.3pt, shadow yshift=-0.5pt}},
        ov/.style={-{Stealth[length=2.4mm,width=2.0mm]}, line width=0.9pt, color=black!70},
        ovp/.style={ov, dash pattern=on 3pt off 2pt},
        aln/.style={{Stealth[length=2.4mm,width=2.0mm]}-{Stealth[length=2.4mm,width=2.0mm]},
            line width=0.9pt, color=black!70, dash pattern=on 3pt off 2pt,
            preaction={draw=white, line width=2.8pt, -}},
        elab/.style={font=\scriptsize\itshape, text=black!55, fill=white, inner sep=1pt},
    }
    \newcommand{\subt}[1]{{\scriptsize\textcolor{black!55}{#1}}}
    \resizebox{0.98\columnwidth}{!}{%
    \begin{tikzpicture}
        \node[cbox, fill=cTrajBg, draw=cTrajFg] (wt) at (0,2.4)
            {\textbf{Wrist Traj.}\\[-1pt]\subt{wrist path $\{\bar{\mathbf{T}}_k\}$}};
        \node[cbox, fill=cPoseBg, draw=cPoseFg] (ps) at (4.6,2.4)
            {\textbf{Ref. Pose}\\[-1pt]\subt{wrist $\bar{\mathbf{T}}$ $+$ fingers $\bar{\mathbf{q}}$}};
        \node[cbox, fill=cHeadBg, draw=cHeadFg] (hd) at (0,0)
            {\textbf{Heading}\\[-1pt]\subt{$\bar{\mathbf{v}},\bar{\mathbf{p}}$ (always on)}};
        \node[cbox, fill=cAffBg, draw=cAffFg] (af) at (4.6,0)
            {\textbf{Affordance}\\[-1pt]\subt{points $\mathcal{O}^{+}/\mathcal{O}^{-}$}};

        \draw[ov]  (hd.north) -- (wt.south);                              %
        \draw[ov]  (af.north) -- (ps.south);                              %
        \draw[ov]  (hd.north east) -- (ps.south west);                    %
        \draw[aln] (af.north west) -- (wt.south east)                     %
            node[elab, midway, above=0pt, sloped] {align};
        \draw[ovp] (hd.east) -- (af.west)                                 %
            node[elab, midway, below=0pt] {implicit};
        \draw[ovp] (ps.west) -- (wt.east)                                 %
            node[elab, midway, above=0pt] {wrist only};

        \draw[-{Stealth[length=2.2mm]}, line width=0.8pt, color=black!45]
            (-1.55,-0.55) -- (-1.55,2.95);
        \node[rotate=90, font=\scriptsize, text=black!55] at (-1.9,1.2)
            {priority: low $\to$ high};

        \draw[-{Stealth[length=2.2mm]}, line width=0.8pt, color=black!45]
            (-0.55,-0.85) -- (5.15,-0.85);
        \node[font=\scriptsize, text=black!55, anchor=north] at (2.3,-0.98)
            {constraint detail: wrist $\to$ full-hand};

        \node[font=\scriptsize, anchor=north] at (2.3,-1.55) {%
            \tikz[baseline=-0.5ex]{\draw[ov] (0,0) -- (0.55,0);}~~override
            \quad
            \tikz[baseline=-0.5ex]{\draw[ovp] (0,0) -- (0.55,0);}~~partial / implicit override
            \quad
            \tikz[baseline=-0.5ex]{\draw[aln] (0,0) -- (0.55,0);}~~alignment%
        };
    \end{tikzpicture}}
    \caption{
    \textbf{Priority among constraints for wrist control.}
    Trajectory and Heading constraints (on the left) specify the wrist motion, while the Reference Pose and Affordance constraints (on the right) specify the wrist pose and finger contact regions. The top two constraints override the bottom two
    as they more explicitly specify the intended motion. 
    }
    \vspace{-2mm}
    \label{fig:constraint_priority}
\end{figure}

%% file: figures/framework.tex
\begin{figure}[t]
    \centering
    \includegraphics[width=\linewidth]{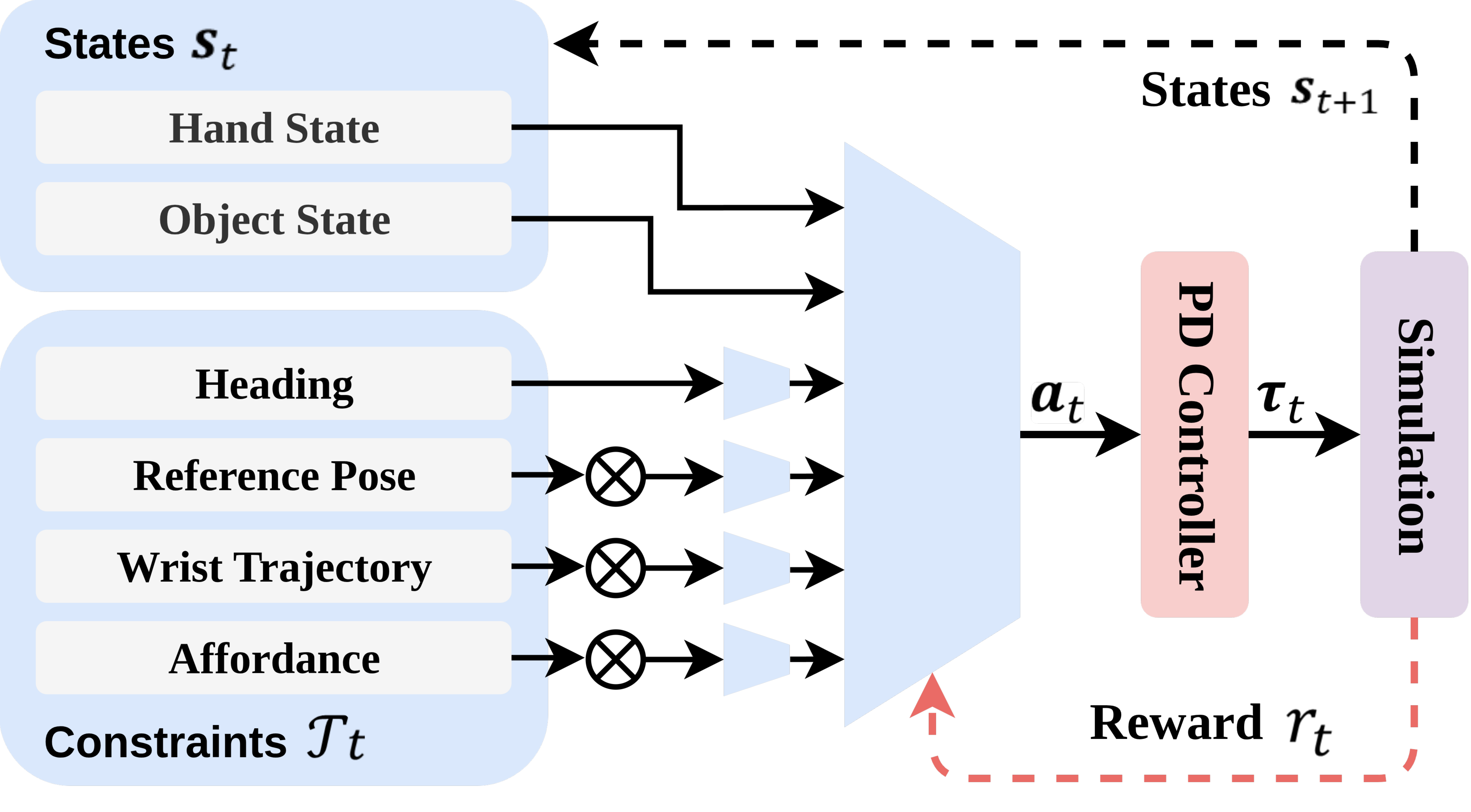}
    \caption{
    \textbf{Network architecture.}
    Hand and object states $\textbf{s}_t$ together with hierarchical constraints $\mathcal{T}_t$ are fed to the main network. The optional constraints are separately encoded and gated by activity masks before concatenation. The policy predicts residual actions $a_t$, which a PD controller executes the corresponding torques $\tau_t$ in simulation to synthesize stable grasping motions that respect the active constraints.
    }
    \label{fig:framework}
\end{figure}

%% file: sec/04_experiment.tex
\section{Experiments}
\label{sec:experiments}

We evaluate whether the policy trained with \method can achieve constraint-conditioned controllability without sacrificing generalization, and demonstrate its capability as a plug-and-play low-level controller for real downstream systems.
Our experiments answer the following questions:
\begin{itemize}
    \item \textbf{Controllability.} Can \method accurately comply with individual constraints and their combinations?
    \item \textbf{Generalization.} Does \method preserve GraspXL-level grasp generalization on large-scale unseen objects and diverse hand morphologies, while transferring its controllability to constraints from unseen datasets?
    \item \textbf{Usability.} With the controllability, can \method serve as a plug-and-play low-level controller that generates physically plausible grasping motions under constraints from different applications?
\end{itemize}

\subsection{Experimental Setup}
\label{sec:exp_setup}

\subsubsection{Implementation Details}
\label{sec:exp_implementation}
The policy is trained with PPO in a physics simulator IsaacGym~\cite{makoviychuk2021isaacgym} which enables GPU-based parallel multi-environment simulation.
Training and evaluation are conducted using a single NVIDIA RTX 4090 GPU. 
Similar to GraspXL, we train the policy in two phases, with the first phase using stationary objects and the second phase using free objects with larger constraint reward weights.
With 4096 environments running in parallel, the training takes around 25 hours in total (15 hours for the first phase and 10 hours for the second).
All the training hyperparameters can be found in the supplementary material.

\subsubsection{Constraint Setup}
\label{sec:exp_constraint_setup}

We extract reference poses and wrist trajectories of the right hand from GRAB~\cite{grab}, OakInk~\cite{yang2022oakink}, and DexYCB~\cite{chao2021dexycb}.
For each sequence, we first extract the grasping frames from GrabNet~\cite{grab} for GRAB, and from the first frame of a five-frame stable contact (thumb plus at least one other finger) for OakInk and DexYCB.
From each grasping frame we search backward along continuous wrist motion to the local maximum of the wrist-object distance, discard sequences whose farthest wrist distance is below $8$\,cm, and truncate the rest to $20$\,cm.
We further remove the reference poses that exceed finger joint anatomical limits (due to noisy captures) and overly complex wrist trajectories (as not all sequences are approach-and-grasp, especially in OakInk), then resample each wrist trajectory to $15$ frames, yielding $244$, $339$, and $92$ sequences for GRAB, OakInk, and DexYCB, respectively.

For OakInk we additionally obtain affordance partitions from it labeled semantic parts.
Specifically, we mark the parts with at least three finger-object contact vertices in the grasping frame as the affordance region $\mathcal{O}^{+}$ and the rest as the non-affordance region $\mathcal{O}^{-}$.
Objects whose affordance region occupies less than $10\%$ of the surface are discarded.

For each object we pre-sample $256$ sets of heading constraints. 
We first sample a point on the affordance surface $\mathcal{O}^{+}$ (or the whole object surface if no affordance is assigned), then sample $\bar{\textbf{v}}$ within a $30^\circ$ cone around the normal at that point, and set $\bar{\textbf{p}}$ by moving the surface point $2$\,cm inward along $\bar{\textbf{v}}$. Candidates that approach $\mathcal{O}^{-}$ (if available) or whose local object width (object point cloud projection along the cross-finger axis $\textbf{w}$) exceeds $12$\,cm are rejected.

\subsubsection{Training Setup}
\label{sec:exp_training_setup}

We extract 80\% of the constraints (reference poses, wrist trajectories, and heading constraints) of GRAB and OakInk for training.
The DexYCB sequences are held out entirely for the cross-dataset evaluation in Sec.~\ref{sec:exp_cross_dataset}.
During training, half of the parallel environments are drawn from GRAB (affordance constraint inactive), and the other half from OakInk (affordance constraint active).
On top of this mix, the wrist-trajectory and reference-pose constraints are independently activated with probability $0.5$ and randomly sampled from the training sets (if active) at each environment reset.
With wrist trajectory and reference pose both inactive, we sample the heading constraint randomly from the training heading pool of the object.

\subsubsection{Hand Initialization}
\label{sec:exp_init}
At each episode reset, the wrist and fingers are initialized from the active constraints.
If a wrist trajectory is provided, the wrist is placed at its first frame \(\bar{\textbf{T}}_{1}\). Otherwise, we align the heading axis \(\textbf{v}\) and the hand center \(\textbf{p}\) with the target heading \(\bar{\textbf{v}}\) and \(\bar{\textbf{p}}\), then displace the wrist \(15\)\,cm along \(-\bar{\textbf{v}}\).
If a reference pose is provided, the fingers are set to the reference joint angles \(\bar{\textbf{q}}\). Otherwise they start from a partially open pose.

\subsubsection{Metrics}
\label{sec:exp_metrics}
We report grasp success rates and constraint-satisfaction errors, averaged over successful episodes with the corresponding constraint active.

\noindent\textbf{Success Rate:}
A grasp is determined as a success if, after the policy executes for $50$ steps, the object does not fall in the following $25$ steps with the table removed.

\noindent\textbf{Heading Error (rad\&m):}
The mean geodesic distance between the heading axis $\textbf{v}$ and the target heading direction $\bar{\textbf{v}}$, measured in radians.
The mean length of the projection of $\textbf{p}-\bar{\textbf{p}}$ onto the plane perpendicular to $\bar{\textbf{v}}$, measured in meters.
Both measured after table removal.

\noindent\textbf{Reference Pose Error (rad\&m):}
The mean absolute joint-angle error over the finger joints $\textbf{q}_t$ and rotational wrist joints $\textbf{u}_{t,4:6}$ relative to $\bar{\textbf{q}}$ and $\bar{\textbf{u}}_{4:6}$, measured in radians. The mean Euclidean distance between the wrist position $\textbf{t}^{w}_t$ and its target $\bar{\textbf{t}}^{w}$, measured in meters.
Both reported after table removal.

\noindent\textbf{Wrist Trajectory Error (rad\&m):}
The mean geodesic and Euclidean distance between the wrist rotation and position with respect to the tracked trajectory waypoint, measured in radians and meters, respectively.
Both measured during approaching before table removal.

\noindent\textbf{Affordance Error (ratio):}
The ratio between the number of links contacted with the non-graspable region $\mathcal{O}^{-}$ and all links contacted with $\mathcal{O}^{+}$ or $\mathcal{O}^{-}$ after table removal.

\input{figures/qualitative_constraints_four_columns}

\subsection{Controllability Evaluation}
\label{sec:exp_controllability}

We first assess the controllability offered by \method, i.e., whether the policy complies with the provided constraints during grasping.
Throughout this evaluation, controllability is only meaningful when coupled with grasp stability: a constraint is useful only if it is satisfied while the grasp remains successful, so we always report constraint-satisfaction errors together with the grasp success rate.
We first study each constraint individually (Sec.~\ref{sec:exp_single}), measuring the grasp success rate and tracking accuracy under a single active constraint and comparing against a dedicated baseline for each control signal. Then we evaluate the performance under combinations of constraints.
The rest 20\% constraints unused during training of GRAB and OakInk are used for this evaluation.
Each setting (each line in the table) is evaluated on 1000 episodes with objects and constraints uniformly distributed.

\input{tables/tab_single_constraint}

\input{figures/noisy_constraints}

\subsubsection{Individual Constraints}
\label{sec:exp_single}

For each constraint type, we provide that constraint alone and measure both its constraint satisfaction error and the resulting grasp success rate.
Heading, reference-pose, and wrist-trajectory constraints are evaluated on GRAB, and the affordance constraint on OakInk. 

\noindent\textbf{Baseline.}
To verify the strong controllability offered by \method, we evaluate and compare its performance under each constraint against a dedicated baseline that targets the same signal: D-Grasp~\cite{christen2022dgrasp} for pose and GraspXL for heading and affordance. As no baseline is available for wrist trajectory, we design a naive baseline with PD control, which first mimics the wrist trajectory frame by frame and then gradually closes fingers. The baseline policies are trained and evaluated in the same dataset as ours with the corresponding constraints for a fair comparison.

\noindent\textbf{Results.}
Quantitative results are shown in Table~\ref{tab:single_constraint}.
Across the four individual constraints, \method consistently achieves high grasp success while maintaining constraint adherence.
Specifically, it reduces angular and positional heading errors by 50.7\% and 28.6\% relative to GraspXL, with a higher success rate.
For reference-pose, \method substantially improves success by 36.2 percentage points over D-Grasp.
While the wrist trajectory tracking error is higher than the PD baseline, which is expected as the PD controller is designed to exactly track the reference trajectory, \method significantly improves success by 31.0 percentage points.
Overall, the results verify the strong controllability with high grasp success of \method.

Fig.~\ref{fig:qualitative_constraints} visualizes the controllability offered by \method.
Conditioned on different constraint settings and configurations, the policy can generate different physically plausible and stable grasping motions.
Notably, the original constraints from the captured datasets are often imperfect.
As illustrated in Fig.~\ref{fig:noisy_constraints}, a reference pose could contain clear finger-object penetration due to noisy captures. Nevertheless, \method can still produce physically stable grasps: rather than strictly imitating the invalid contact, the policy adapts the fingers during interaction and forms stable contacts.
This robustness to noisy upstream signals is important when \method is used as a plug-and-play low-level controller.

\input{tables/tab_constraint_composition}

\subsubsection{Composed Constraints}
\label{sec:exp_composition}

We further evaluate whether \method can preserve grasp success and constraint satisfaction when multiple constraints are jointly specified.
We enumerate combinations that can co-exist without overriding each other under the priority hierarchy of Sec.~\ref{sec:constraint}, i.e., wrist trajectory with reference pose (Ref. + Traj., on GRAB) and wrist trajectory with affordance (Traj. + Aff., on OakInk).

As shown in Table~\ref{tab:constraint_composition}, combining multiple constraints leads to some performance degradation relative to the corresponding single-constraint settings in Table~\ref{tab:single_constraint}, reflecting the difficulty of satisfying potentially misaligned constraints simultaneously. Nevertheless, the degradation remains modest across most metrics. The largest performance gap occurs in affordance error, likely because prescribed wrist trajectories can bring the hand close to the boundary between affordance and non-affordance regions, increasing incidental finger contact with non-affordance regions during trajectory tracking.

\subsection{Generalization Evaluation}
\label{sec:exp_generalization}
In this section, we test whether \method can generalize its controllability on new datasets, and preserve the GraspXL-level generalizable grasping performance on large-scale unseen objects and diverse hand morphologies.

\input{tables/tab_cross_dataset_controllability}

\subsubsection{Cross-Dataset Controllability}
\label{sec:exp_cross_dataset}
To verify the constraint-conditioned controllability transfers beyond the training distribution, we further evaluate the trained policy on DexYCB~\cite{chao2021dexycb}, where both the objects and human-captured constraints are unseen during training. As it does not contain affordance constraints, we only evaluate the heading, wrist trajectory, and reference pose constraints, and their combinations.
Each constraint and their combination is evaluated on 1000 episodes with objects and constraints uniformly distributed.

The results are shown in Table~\ref{tab:cross_dataset_controllability}.
Performance under heading constraints remains highly comparable to that on GRAB, indicating that the policy generalizes well to unseen objects.
Success rates decrease under reference pose and wrist trajectory constraints, which we attribute to differences in motion patterns between the datasets.
For example, DexYCB sequences begin with the hand below the tabletop, followed by an arm lift above the object before approaching and grasping it, whereas GRAB features more natural approach-to-grasp motions.
These distinct motion patterns introduce a domain gap in the human-captured constraints.
Despite the reduced success rates, constraint errors remain broadly comparable, with both improvements and degradations across metrics, supporting the cross-dataset generalization of controllability.

\subsubsection{Large-Scale Object Geometry}
\label{sec:exp_object}
To show \method retains the generalization of GraspXL on large-scale objects, we evaluate our trained policy on the 500k+ Objaverse~\cite{objaverse} objects used in GraspXL and report success rates stratified by object size.
During evaluation, no task-specific reference pose, wrist trajectory, or affordance region is provided, with the heading constraint sampled randomly as in Sec.~\ref{sec:exp_constraint_setup}. 

The results are shown in Table~\ref{tab:main_generalization}.
\method outperforms GraspXL across all object size categories, with the largest gain on large objects, yielding more consistent performance across sizes.
The dynamic hand center contributes to this improvement by its adaptation as the hand opens and closes for different object sizes, leading the hand to effectively enclose objects as illustrated in Fig.~\ref{fig:dynamic_hand_center}.
To verify this contribution, we replace the dynamic hand center with the original fixed GraspXL center for comparison, and the clear performance drop as shown in the table (w/o Dyn. Ctr.) supports the claim.
Overall, these results demonstrate that \method maintains and further improves the large-scale generalization capability of GraspXL.

\input{tables/tab_main_generalization}

\subsubsection{Hand Morphology}
Our embodiment-agnostic framework supports constraint-conditioned grasp synthesis across different hand morphologies.
We evaluate this capability on MANO~\cite{MANO:SIGGRAPHASIA:2017}, Allegro~\cite{Allegro}, and Sharpa~\cite{Sharpa} under a unified protocol, scaling objects by $1.4\times$ for Allegro and $1.1\times$ for Sharpa to accommodate the larger robot hands.
Heading and affordance constraints follow the same sampling as in Sec.~\ref{sec:exp_constraint_setup} as they are defined on the object and thus transfer across hand morphologies without modification.
Since comparable reference poses and wrist trajectories are unavailable across these hand models, we procedurally generate them for each hand including MANO.

Specifically, we first train a policy for each hand without reference-pose and wrist-trajectory constraints, extract the final states of successful grasps as reference poses. 
We then reverse-sample wrist trajectories from these poses by moving the wrist outward by $15$--$25$\,cm along a direction sampled within a $15^\circ$ cone around the heading, adding a lateral sinusoidal displacement with an amplitude of $1$--$5$\,cm and a random wrist rotation within $30^\circ$.
Each path is replayed using the wrist PD controller, and the wrist pose actually reached at each timestep is recorded.
We generate 20 reference poses and 20 wrist trajectories per object and hand.

For each hand, all constraint sets are split into 80\% for training and 20\% for evaluation.
We then train a separate constraint-conditioned policy for each hand, with all other experimental settings following Sec.~\ref{sec:exp_single}. 

The results are shown in Table~\ref{tab:hand_generalization}.
Overall, \method maintains high grasp success with low constraint errors across all three morphologies.
Compared with Table~\ref{tab:single_constraint}, MANO exhibits substantially lower reference-pose and wrist-trajectory errors, with reference-pose success also increasing from 87.4\% to 96.4\%.
We attribute these improvements to the more consistent motion patterns of the constraints and the absence of capture noise in the generated reference poses and wrist trajectories.
Among the three hands, Sharpa consistently achieves the highest success rates, with the errors being lowest on most metrics, benefiting from its agile morphology design.
Conversely, Allegro's larger palm and fewer degrees of freedom limits its dexterity and causes relatively lower success rates, while its samller number of finger links also lead to less unintended contact outside the target affordance region, yielding the lowest affordance contact error. 
Together, these results show that our framework consistently provides constraint-conditioned controllability across diverse hand morphologies.

\input{tables/tab_hand_generalization}

\subsection{Ablation}
\label{sec:exp_ablation}
Beyond evaluating the dynamic hand center for large-scale object generalization in Table~\ref{tab:main_generalization}, we examine how the dynamic hand center and feed-forward guidance affect controllability under the single-constraint settings of Table~\ref{tab:single_constraint}.
Specifically, we replace the dynamic hand center with the fixed GraspXL midpoint (w/o Dyn. Ctr.) and remove the one-step lookahead from the guidance target by using $\bar{\mathbf{T}}_{k^{*}}$ instead of $\bar{\mathbf{T}}_{k^{*}+1}$ (w/o FF Guid.).
Since the other single-constraint settings do not use this lookahead and are therefore unaffected by its removal, we report w/o FF Guid. only for wrist-trajectory control.

As shown in Table~\ref{tab:ablation}, removing the dynamic hand center reduces success rates across all constraint settings, with the largest drop under wrist-trajectory control. It also increases errors for all constraints except affordance. These results further highlight the importance of the dynamic hand center for consistent successful grasping and accurate constraint satisfaction. 
Removing the one-step lookahead also substantially reduces success under wrist-trajectory constraints, indicating that it supports more reliable grasp completion by encouraging forward progress along instead of just mimicing the trajectory.

\input{tables/tab_ablation}

\input{figures/qualitative_applications}

\subsection{Usability Examples}
\label{sec:exp_usability}

Finally, we qualitatively demonstrate several direct applications with the policy serving as a low-level controller.

\subsubsection{Functional Grasping}
As shown in the second column of Fig.~\ref{fig:teaser}, given a functional reference grasping pose, \method can effectively follow the reference fingers and wrist to reproduce a task-appropriate grasp on the target object while keeping it physically stable.

\subsubsection{Whole-Body Grasp Completion}
As shown in the fourth column of Fig.~\ref{fig:teaser}, given a wrist trajectory from a whole-body motion generation system, \method can synthesize finger motions that establish accurate contact with the local object geometry while tracking the wrist trajectory, bridging high-level body motion and contact-rich finger control.

\subsubsection{Human-Motion Imitation}
As shown in Fig.~\ref{fig:qualitative_applications}, \method can utilize the task-driven constraints extracted from a human grasping motion to synthesize corresponding grasping motions for different dexterous robot hands, providing an effective and efficient way to transfer human grasping skills to robot hands.

%% file: figures/qualitative_constraints_four_columns.tex
\newlength{\qualcolcellw}
\newlength{\qualcolinputw}
\newlength{\qualcolseqw}
\newlength{\qualcolgap}
\newlength{\qualcolrowh}
\newlength{\qualcolframeh}

\newcommand{\qualcolpairedrow}[3]{%
    \setlength{\qualcolcellw}{#1}%
    \setlength{\qualcolinputw}{0.30\qualcolcellw}%
    \setlength{\qualcolgap}{0.025\qualcolcellw}%
    \setlength{\qualcolseqw}{0.675\qualcolcellw}%
    \setlength{\qualcolrowh}{0.50\qualcolcellw}%
    \setlength{\qualcolframeh}{0.34\qualcolcellw}%
    \begin{tikzpicture}[baseline=(current bounding box.south)]
        \path[use as bounding box] (0,0)
            rectangle (\qualcolcellw,\qualcolrowh);
        \node[anchor=south west, inner sep=0,
              minimum width=\qualcolinputw,
              minimum height=\qualcolrowh] (inputspace) at (0,0) {%
            \begin{tikzpicture}
                \node[draw=black!30, line width=0.4pt, rounded corners=2pt,
                      inner sep=3pt, minimum width=\qualcolinputw,
                      minimum height=\qualcolframeh] {%
                    \includegraphics[width=\dimexpr\qualcolinputw-6pt\relax,
                                     height=\dimexpr\qualcolframeh-6pt\relax,
                                     keepaspectratio]{#2}};
            \end{tikzpicture}};
        \node[anchor=south west, inner sep=0,
              minimum width=\qualcolseqw,
              minimum height=\qualcolrowh] at
            ([xshift=\qualcolgap]inputspace.south east) {%
            \includegraphics[width=\qualcolseqw,
                             height=\qualcolrowh,
                             keepaspectratio]{#3}};
    \end{tikzpicture}%
}

\newcommand{\qualcolpanel}[3]{%
    \begin{minipage}[b]{0.96\linewidth}
        \centering
        \qualcolpairedrow{\linewidth}{#11_#2}{#11_#3}\\[3pt]
        \qualcolpairedrow{\linewidth}{#12_#2}{#12_#3}\\[3pt]
        \qualcolpairedrow{\linewidth}{#13_#2}{#13_#3}
    \end{minipage}%
}

\begin{figure*}[t]
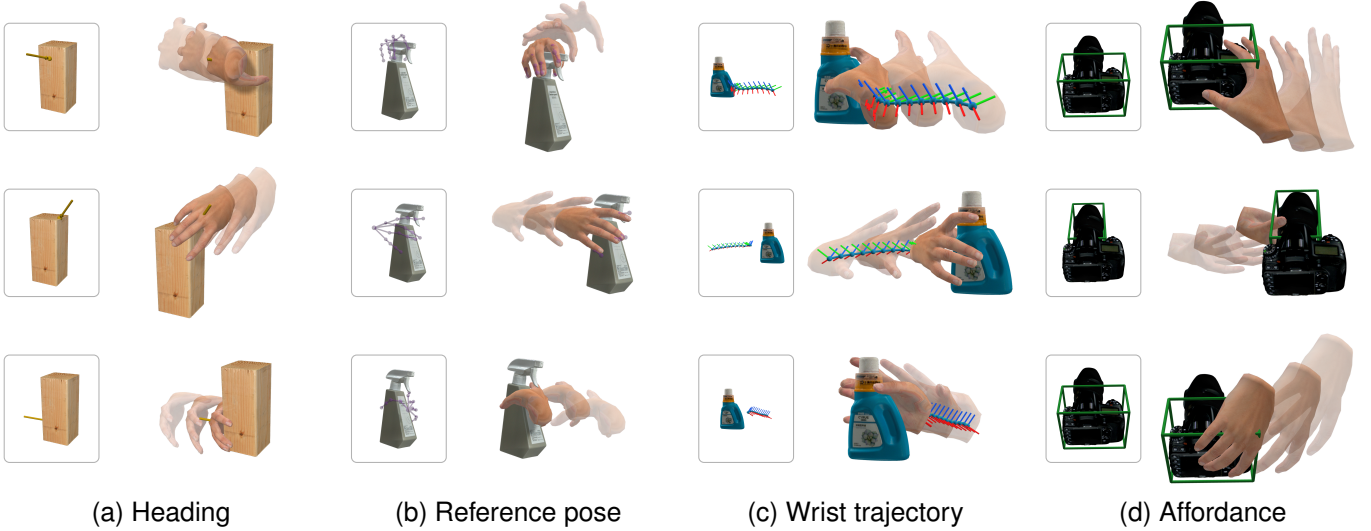

    \centering
    \begin{minipage}[b]{0.24\linewidth}
        \centering
        \subfloat[Heading]{%
            \qualcolpanel{FIG/lossless/quat_new/head_row}{constraint}{seq}}
    \end{minipage}
    \hfill
    \begin{minipage}[b]{0.24\linewidth}
        \centering
        \subfloat[Reference pose]{%
            \qualcolpanel{FIG/lossless/quat_new/row}{ref}{seq}}
    \end{minipage}
    \hfill
    \begin{minipage}[b]{0.24\linewidth}
        \centering
        \subfloat[Wrist trajectory]{%
            \qualcolpanel{FIG/lossless/quat_new/traj_row}{constraint}{seq}}
    \end{minipage}
    \hfill
    \begin{minipage}[b]{0.24\linewidth}
        \centering
        \subfloat[Affordance]{%
            \qualcolpanel{FIG/lossless/quat_new/aff_row}{constraint}{seq}}
    \end{minipage}
    \caption{
    \textbf{Qualitative examples under different constraints.}
    The same policy generates different physically plausible grasping motions when conditioned on different heading, reference pose, wrist trajectory, and contact constraints. Each column corresponds to one constraint type and shows three grasping sequences.
    }
    \label{fig:qualitative_constraints}
\end{figure*}

%% file: tables/tab_single_constraint.tex
\begin{table}[t]
    \centering
    \caption{
    Controllability Evaluation on Single Constraint.
    }
    \label{tab:single_constraint}
    \resizebox{\linewidth}{!}{
    \begin{tabular}{l | l | c c}
        \toprule
        Constraint & Method & Success $\uparrow$ & Constraint Error $\downarrow$ \\
        \midrule
        \multirow{2}{*}{Heading} & GraspXL~\cite{zhang2024graspxl} & 94.2\% & 0.278\,rad / 0.028\,m \\
         & \method (Ours) & 96.9\% & 0.137\,rad / 0.020\,m \\ %
        \midrule
        \multirow{2}{*}{Ref. Pose} & D-Grasp~\cite{christen2022dgrasp} & 51.2\% & 0.390\,rad / 0.031\,m \\
         & \method (Ours) & 87.4\% & 0.316\,rad / 0.036\,m \\
        \midrule
        \multirow{2}{*}{Wrist Traj.} & PD & 64.5\% & 0.160\,rad / 0.019\,m \\
         & \method (Ours) & 95.5\% & 0.162\,rad / 0.029\,m \\
        \midrule
        \multirow{2}{*}{Afford.} & GraspXL~\cite{zhang2024graspxl} & 91.9\% & 0.016 \\
         & \method (Ours) & 98.2\% & 0.016 \\
        \bottomrule
    \end{tabular}
    }
\end{table}

%% file: figures/noisy_constraints.tex
\begin{figure}[t]
    \centering
    \begin{tikzpicture}[baseline=(current bounding box.center)]
        \pgfmathsetmacro{\leftH}{0.175}
        \pgfmathsetmacro{\leftW}{\leftH*3988/3240}
        \pgfmathsetmacro{\gapfrac}{22pt/\linewidth}
        \pgfmathsetmacro{\seqfrac}{1-\leftW-\gapfrac-10pt/\linewidth}
        \pgfmathsetmacro{\nOneW}{\seqfrac*1733/4453}
        \pgfmathsetmacro{\nTwoW}{\seqfrac*1430/4453}
        \pgfmathsetmacro{\nThreeW}{\seqfrac*1290/4453}

        \pgfmathsetmacro{\nTwoDy}{108*\nTwoW/1430 - 121*\nOneW/1733}
        \pgfmathsetmacro{\nThreeDy}{738*\nThreeW/2220 - 121*\nOneW/1733}

        \node[inner sep=0] (L) {%
            \includegraphics[height=\leftH\linewidth]{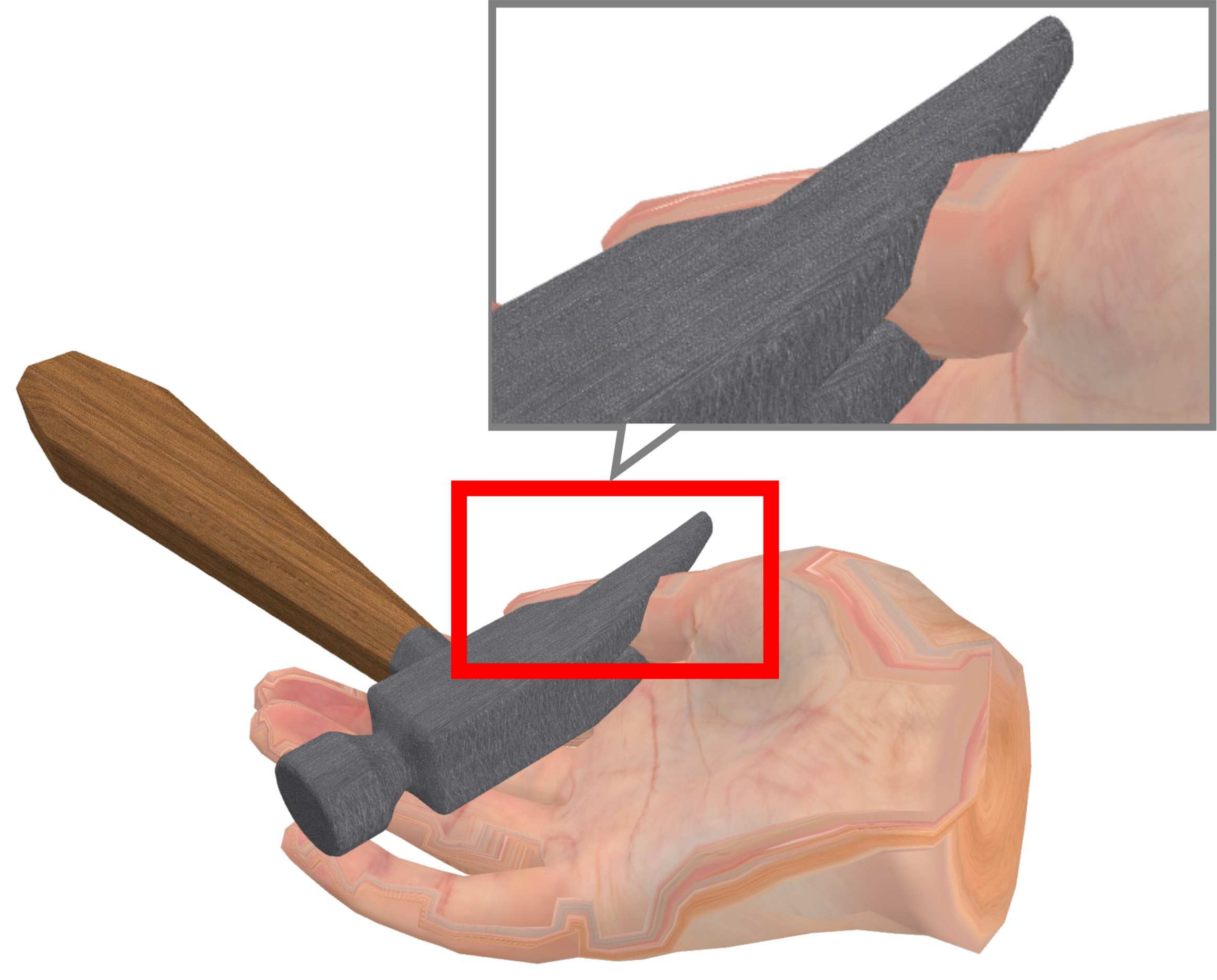}};
        \node[inner sep=0, anchor=west] (N1) at ([xshift=14pt]L.east) {%
            \includegraphics[width=\nOneW\linewidth]{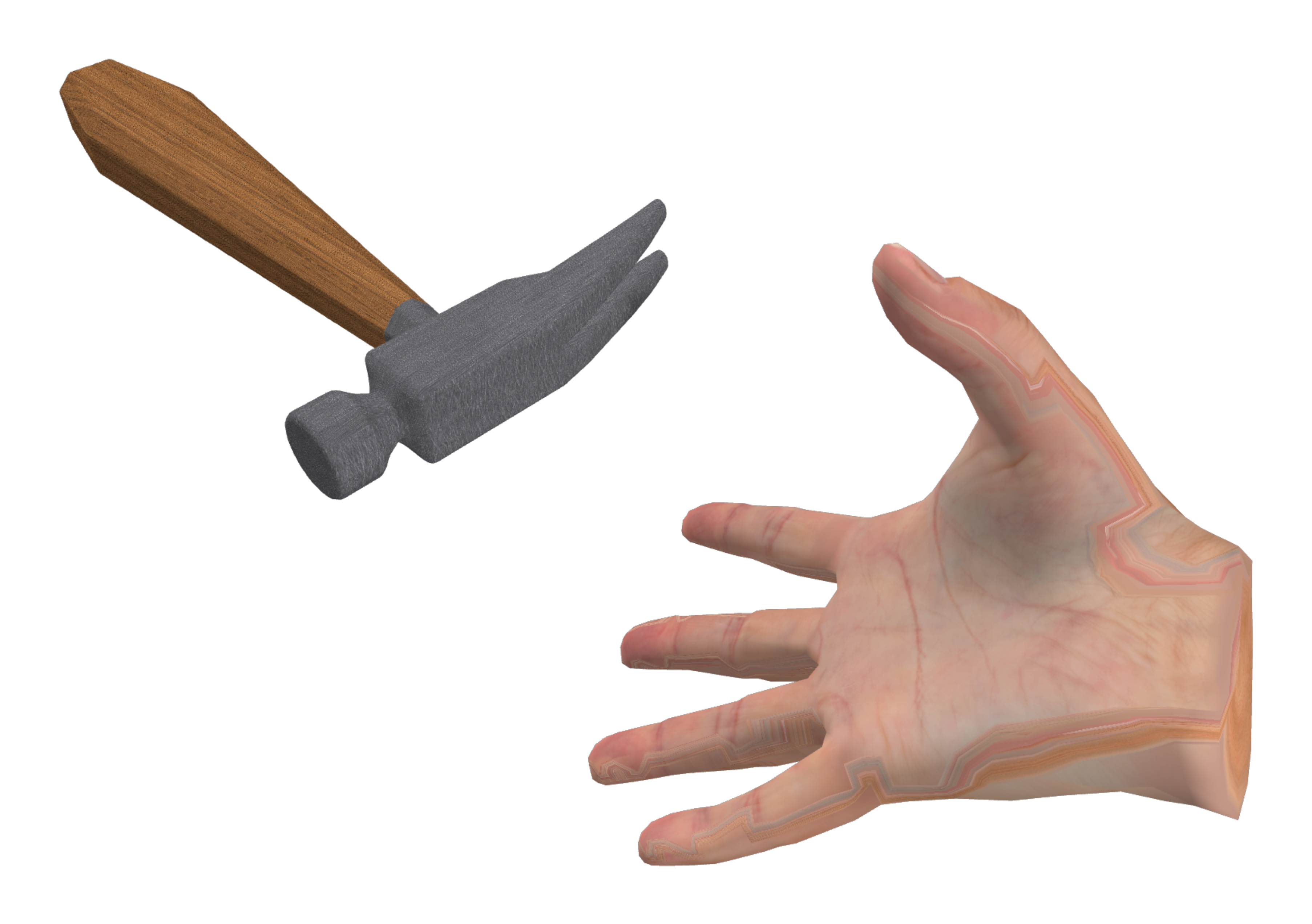}};
        \node[inner sep=0, anchor=north west] (N2) at
            ([xshift=4pt, yshift=\nTwoDy\linewidth]N1.north east) {%
            \includegraphics[width=\nTwoW\linewidth]{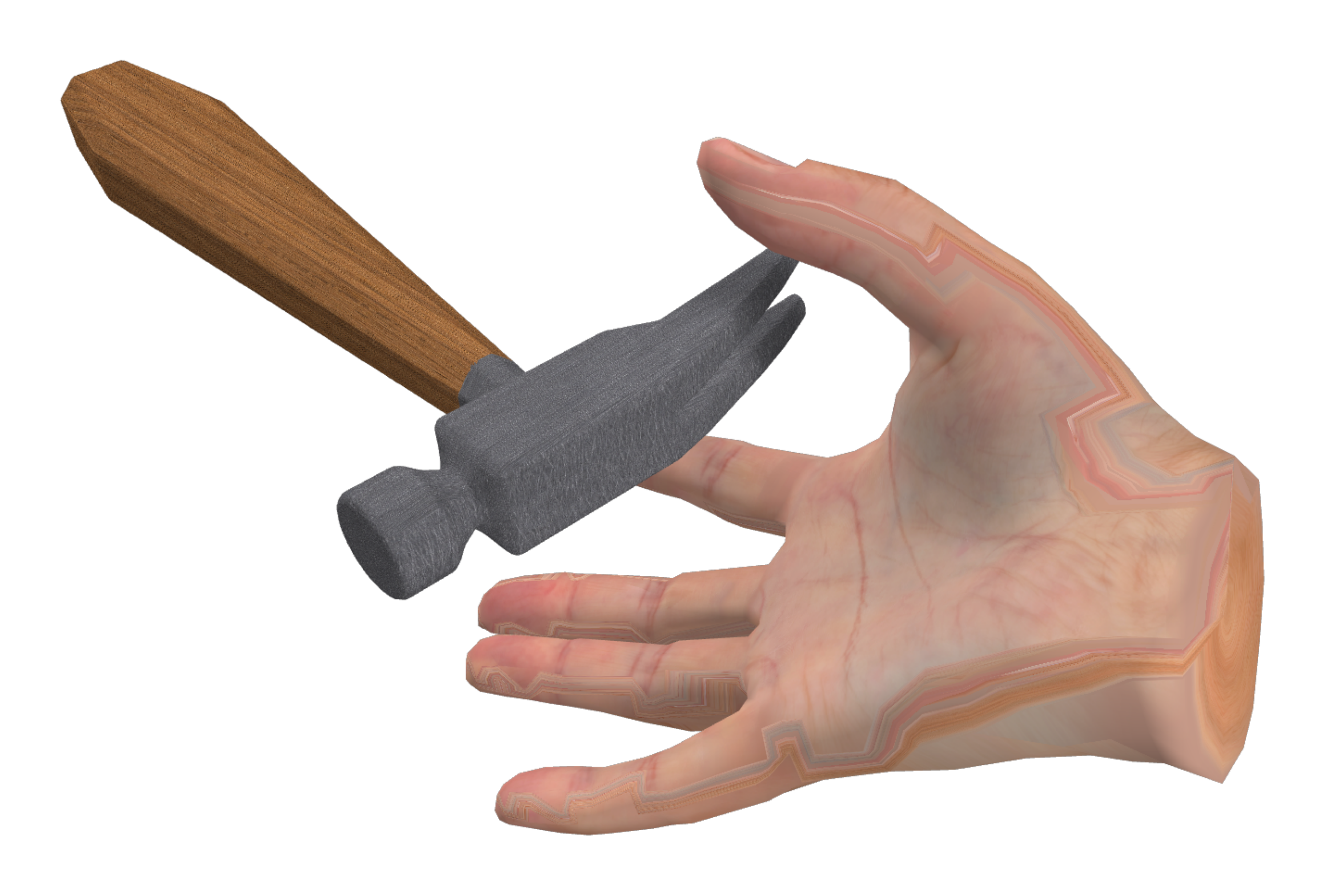}};
        \node[inner sep=0, anchor=north west] (N3) at
            ([xshift=4pt, yshift=\nThreeDy\linewidth]N2.east |- N1.north) {%
            \includegraphics[width=\nThreeW\linewidth]{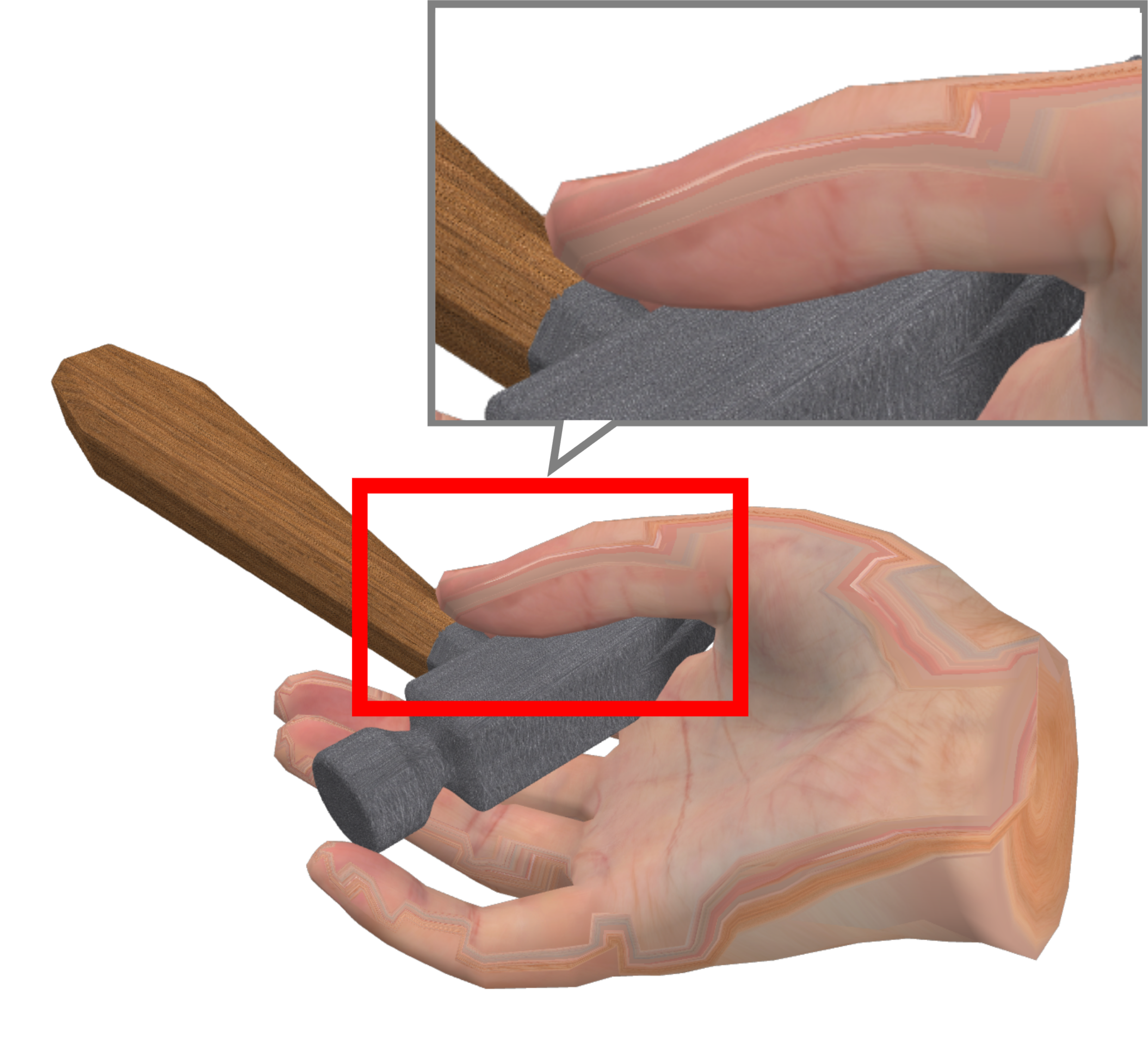}};
        \node[fit=(N1)(N2)(N3), inner sep=0] (R) {};
        \coordinate (M) at ($(L.east)!0.5!(N1.west)$);
        \draw[dashed, line width=0.7pt, black!45]
            ([yshift=3pt]M |- R.south) -- ([yshift=-3pt]M |- R.north);
        \coordinate (T1) at ($(N1.west |- R.south)+(2pt,-3pt)$);
        \coordinate (T3) at ($(N3.east |- R.south)+(-2pt,-3pt)$);
        \path[shade, left color=black!10, right color=black!48]
            ($(T1)+(0,1.5pt)$) -- ($(T3)+(-4pt,1.5pt)$) --
            ($(T3)+(-4pt,3.2pt)$) -- (T3) --
            ($(T3)+(-4pt,-3.2pt)$) -- ($(T3)+(-4pt,-1.5pt)$) --
            ($(T1)+(0,-1.5pt)$) -- cycle;
        \node[fill=white, font=\scriptsize\sffamily, text=black!55,
              inner xsep=2.5pt, inner ysep=0]
            at ($(T1)!0.5!(T3)$) {time};
        \node[anchor=north, font=\small\sffamily, inner sep=0]
            at ([yshift=-11pt]L.center |- R.south) {Noisy reference};
        \node[anchor=north, font=\small\sffamily, inner sep=0]
            at ([yshift=-11pt]R.center |- R.south) {Generated grasping motion};
    \end{tikzpicture}
    \vspace{3pt}
    \caption{
    \textbf{Robustness under noisy constraints.}
    Given a noisy reference pose with finger-object penetration (left), our policy still produces a physically plausible grasping motion with stable contacts (right).
    }
    \label{fig:noisy_constraints}
\end{figure}
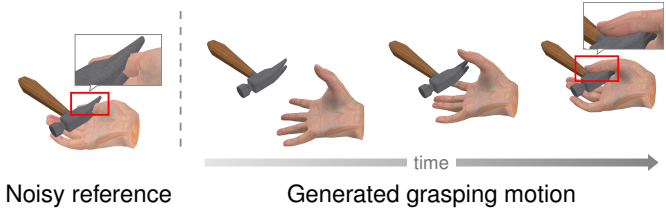

%% file: tables/tab_constraint_composition.tex
\begin{table}[t]
    \centering
    \caption{
    Controllability Evaluation on Composed Constraints.
    }
    \label{tab:constraint_composition}
    \resizebox{\linewidth}{!}{
    \begin{tabular}{l | c |c c c}
        \toprule
        \multirow{2}{*}[-0.5ex]{Constraint Set} & Suc. $\uparrow$ & Pose E. $\downarrow$ & Traj. E. $\downarrow$ & Aff. E. $\downarrow$ \\
         & (\%) & (rad/m) & (rad/m) & (ratio) \\
        \midrule
        Ref. + Traj. & 86.5 & 0.327 / 0.046 & 0.156 / 0.028 & -- \\
        Traj. + Aff. & 92.3 & -- & 0.185 / 0.027 & 0.045 \\
        \bottomrule
    \end{tabular}
    }
\end{table}

%% file: tables/tab_cross_dataset_controllability.tex
\begin{table}[t]
    \centering
    \caption{
    Cross-dataset controllability generalization on DexYCB.
    }
    \label{tab:cross_dataset_controllability}
    \resizebox{\linewidth}{!}{
    \begin{tabular}{l | c | c c c}
        \toprule
        \multirow{2}{*}[-0.5ex]{Constraint} & Suc. $\uparrow$ & Head. E. $\downarrow$ & Pose E. $\downarrow$ & Traj. E. $\downarrow$ \\
         & (\%) & (rad/m) & (rad/m) & (rad/m) \\
        \midrule
        Heading & 97.6 & 0.148 / 0.021 & -- & -- \\
        Ref. Pose & 81.7 & -- & 0.270 / 0.036 & -- \\
        Wrist traj. & 73.3 & -- & -- & 0.183 / 0.029 \\
        Ref. + Traj. & 69.8 & -- & 0.268 / 0.048 & 0.177 / 0.028 \\
        \bottomrule
    \end{tabular}
    }
    \vspace{-1mm}
\end{table}

%% file: tables/tab_main_generalization.tex
\begin{table}[t]
    \centering
    \caption{
    Large-scale object geometry generalization on Objaverse.
    }
    \label{tab:main_generalization}
    \resizebox{\linewidth}{!}{
    \begin{tabular}{l | c c c | c}
        \toprule
        Method & Small & Medium & Large & Average \\
        \midrule
        GraspXL~\cite{zhang2024graspxl} & 85.9\% & 84.5\% & 79.0\% & 82.2\% \\
        \method (Ours) & 92.9\% & 93.6\% & 91.7\% & 92.7\% \\ %
        Ours (w/o Dyn. Ctr.) & 87.1\% & 89.3\% & 87.7\% & 88.0\% \\
        \bottomrule
    \end{tabular}
    }
\end{table}

%% file: tables/tab_hand_generalization.tex
\begin{table}[t]
    \centering
    \caption{
    Hand morphology generalization.
    }
    \label{tab:hand_generalization}
    \resizebox{\linewidth}{!}{
    \begin{tabular}{l | l | c c}
        \toprule
        Constraint & Hand & Success $\uparrow$ & Constraint Error $\downarrow$ \\
        \midrule
        \multirow{3}{*}{Heading} & MANO~\cite{MANO:SIGGRAPHASIA:2017} & 96.0\% & 0.143\,rad / 0.020\,m \\
        & Allegro~\cite{Allegro} & 90.4\% & 0.160\,rad / 0.016\,m \\ 
        & Sharpa~\cite{Sharpa} & 96.7\% & 0.114\,rad / 0.013\,m \\
       \midrule
       \multirow{3}{*}{Ref. Pose} & MANO~\cite{MANO:SIGGRAPHASIA:2017} & 96.4\% & 0.253\,rad / 0.017\,m \\
        & Allegro~\cite{Allegro}  & 95.7\% & 0.145\,rad / 0.019\,m \\ 
        & Sharpa~\cite{Sharpa} & 99.2\% & 0.145\,rad / 0.014\,m \\
       \midrule
       \multirow{3}{*}{Wrist Traj.} & MANO~\cite{MANO:SIGGRAPHASIA:2017} & 95.0\% & 0.082\,rad / 0.020\,m \\
        & Allegro~\cite{Allegro}  & 94.4\% & 0.082\,rad / 0.026\,m \\ 
        & Sharpa~\cite{Sharpa} & 98.7\% & 0.079\,rad / 0.022\,m \\
       \midrule
       \multirow{3}{*}{Afford.} & MANO~\cite{MANO:SIGGRAPHASIA:2017} & 95.6\% & 0.014 \\
        & Allegro~\cite{Allegro}  & 91.8\% & 0.006 \\
        & Sharpa~\cite{Sharpa} & 95.9\% & 0.010 \\
       \bottomrule
    \end{tabular}
    }
\end{table}

%% file: tables/tab_ablation.tex
\begin{table}[t]
    \centering
    \caption{
    Ablation study under single-constraint controllability.
    }
    \label{tab:ablation}
    \resizebox{0.99\linewidth}{!}{
    \begin{tabular}{l | l | c c}
        \toprule
        Constraint & Variant & Success $\uparrow$ & Constraint Error $\downarrow$ \\
        \midrule
        \multirow{2}{*}{Heading} & Full  & 96.9\% & 0.137\,rad / 0.020\,m \\
         & w/o Dyn. Ctr. & 89.4\% & 0.182\,rad / 0.021\,m \\
        \midrule
        \multirow{2}{*}{Ref. Pose} & Full & 87.4\% & 0.316\,rad / 0.036\,m \\
         & w/o Dyn. Ctr. & 76.0\% & 0.352\,rad / 0.041\,m \\
        \midrule
        \multirow{3}{*}{Wrist Traj.} & Full & 95.5\% & 0.162\,rad / 0.029\,m \\
         & w/o Dyn. Ctr. & 66.7\% & 0.169\,rad / 0.033\,m \\
         & w/o FF Guid. & 76.0\% & 0.170\,rad / 0.029\,m \\
        \midrule
        \multirow{2}{*}{Afford.} & Full & 98.2\% & 0.016 \\
         & w/o Dyn. Ctr. & 93.1\% & 0.013 \\
        \bottomrule
    \end{tabular}
    }
\end{table}

%% file: figures/qualitative_applications.tex
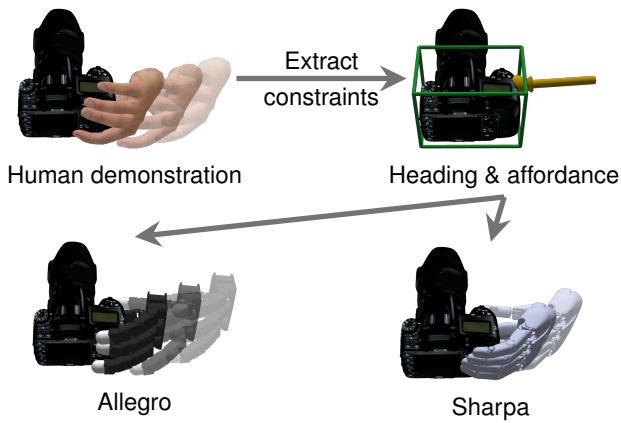
\begin{figure}[t]
    \centering
    \begin{tikzpicture}[
        font=\small\sffamily,
        flow/.style={-{Stealth[length=3.5mm,width=3.4mm]},
                     draw=black!55, line width=2pt},
        image/.style={inner sep=0pt, outer sep=0pt},
        label/.style={inner sep=0pt, align=center}
    ]
        \node[image, anchor=north west] (mano) at (0.04\linewidth,0) {%
            \includegraphics[height=0.225\linewidth]{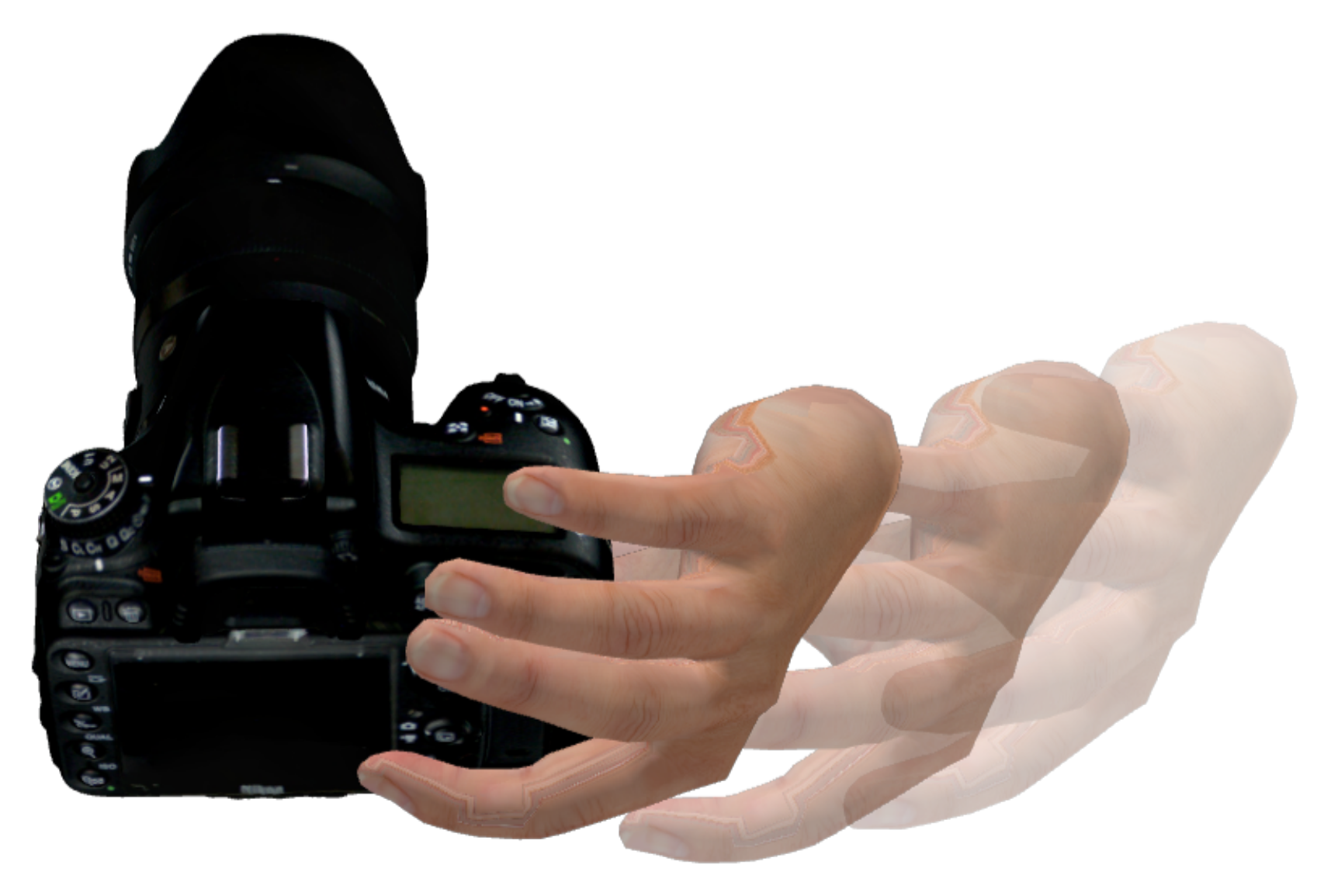}};
        \node[image, anchor=north east] (constraints) at (0.92\linewidth,0) {%
            \includegraphics[height=0.225\linewidth]{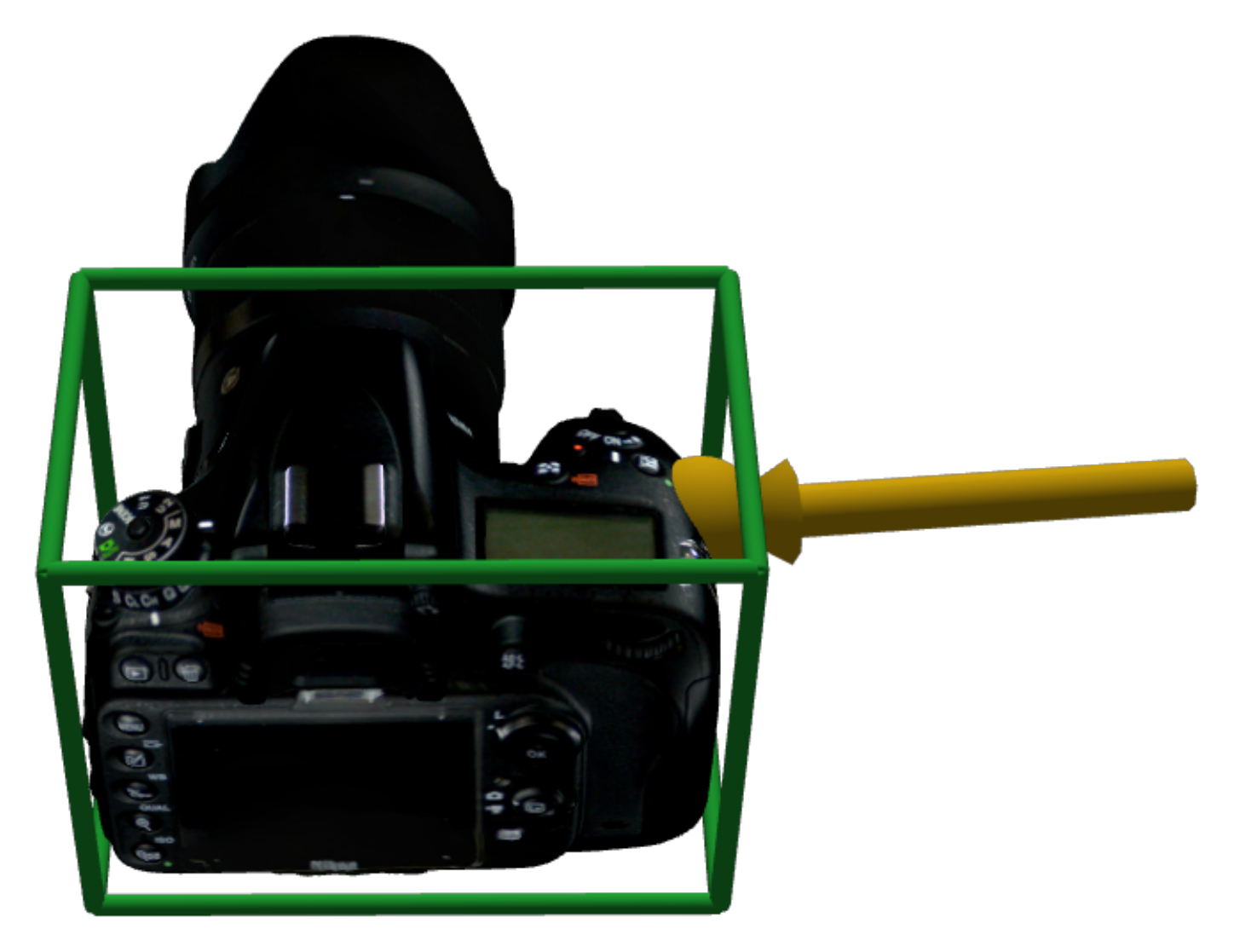}};

        \node[label, below=1.5mm of mano] (mano-label) {Human demonstration};
        \node[label, below=1.5mm of constraints] (constraints-label)
            {Heading \& affordance};

        \draw[flow] (mano.east) --
            node[midway, above=2.5pt, inner sep=0pt] {Extract}
            node[midway, below=2.5pt, inner sep=0pt] {constraints}
            (constraints.west);

        \node[image, below=7mm of mano-label, xshift=0.015\linewidth] (allegro) {%
            \includegraphics[height=0.215\linewidth]{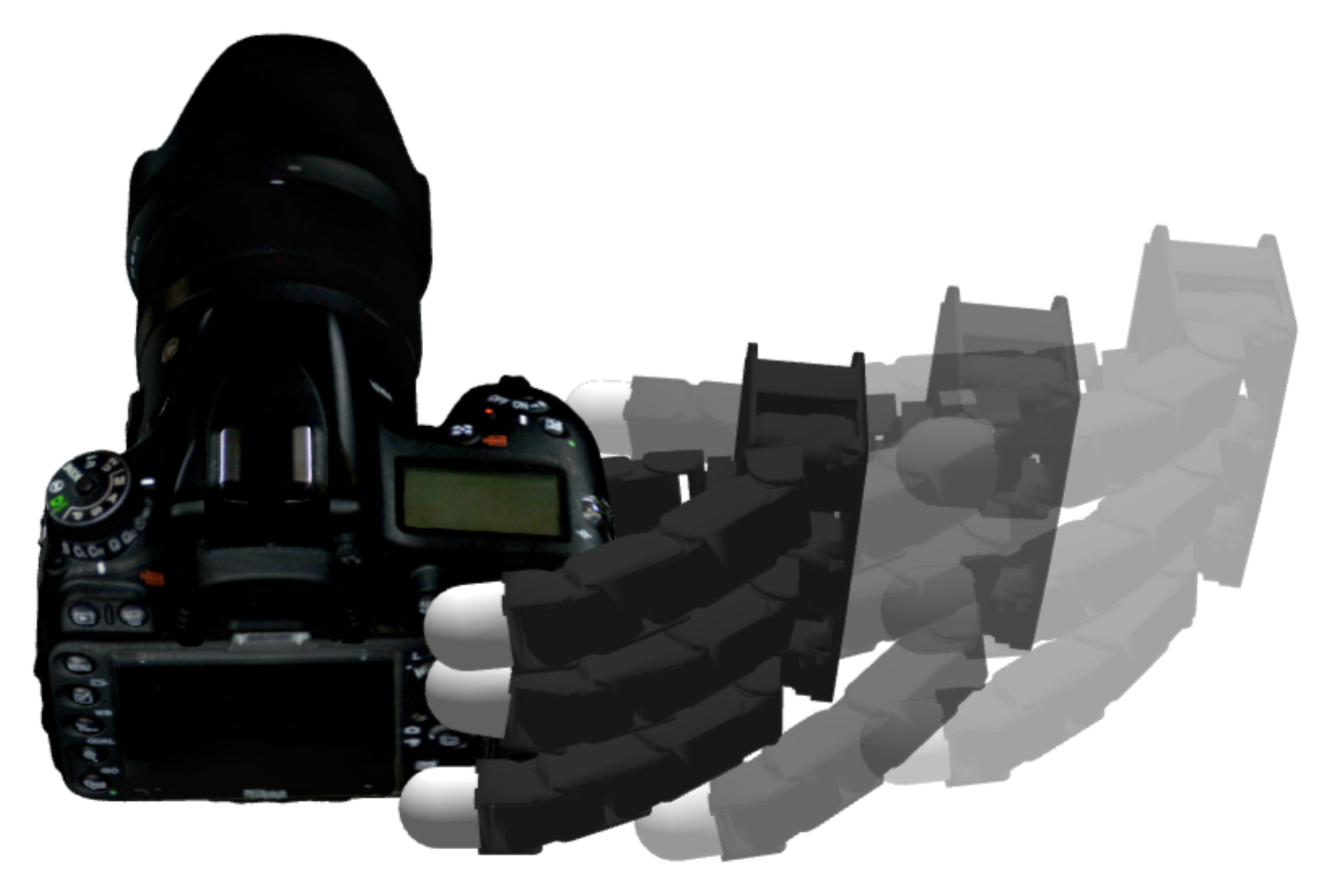}};
        \node[image, below=7mm of constraints-label, xshift=-0.02\linewidth] (sharpa) {%
            \includegraphics[height=0.215\linewidth]{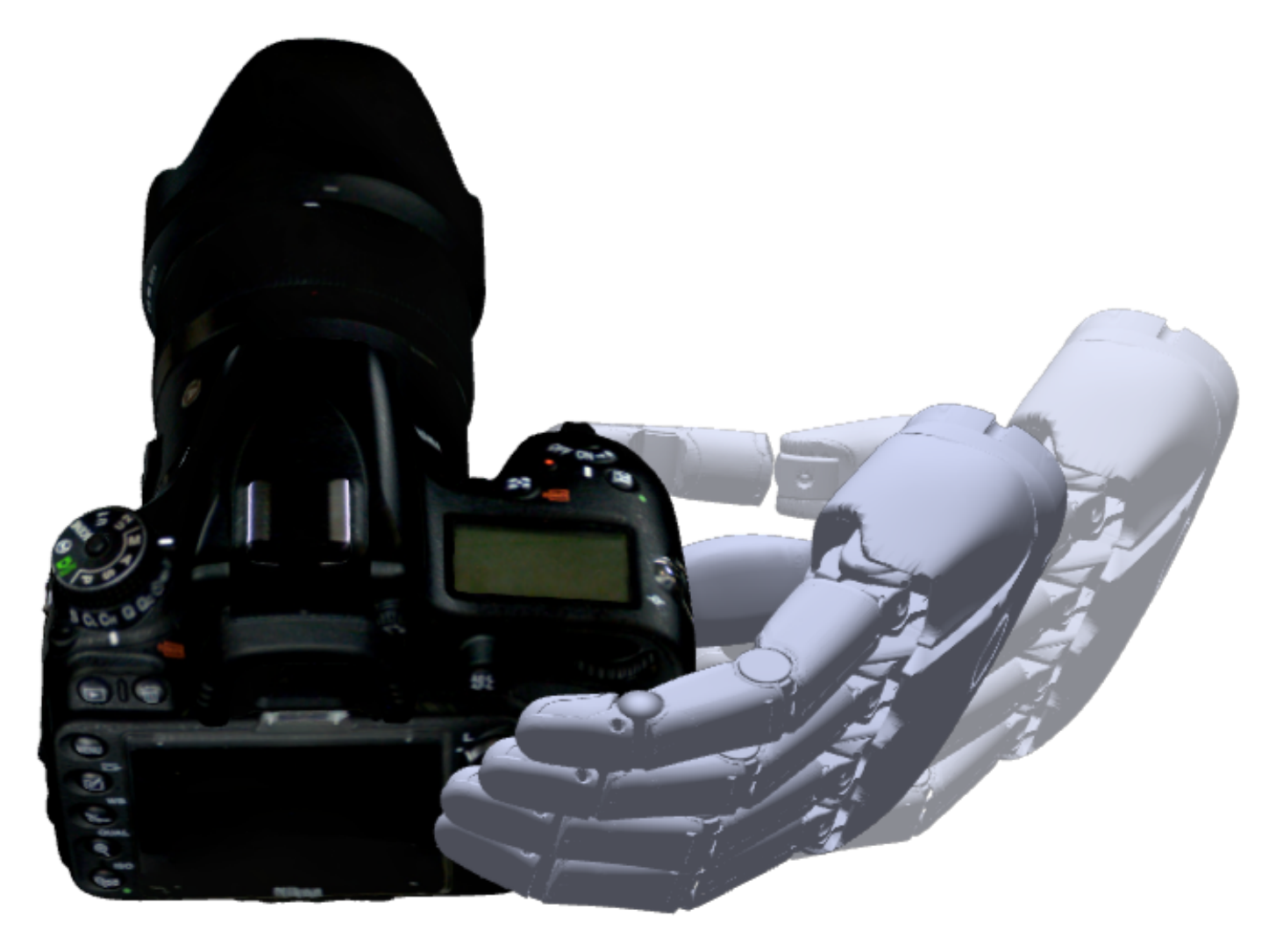}};

        \draw[flow] ([yshift=-3pt]constraints-label.south) -- (allegro.north);
        \draw[flow] ([yshift=-3pt]constraints-label.south) -- (sharpa.north);

        \node[label, below=1.5mm of allegro] {Allegro};
        \node[label, below=1.5mm of sharpa] {Sharpa};
    \end{tikzpicture}
    \caption{
    \textbf{Cross-embodiment motion synthesis from a human demonstration.}
    \method can be used to synthesize corresponding grasping motions for different dexterous robot hands utilizing the task-driven constraints extracted from a human demonstration.
    }
    \label{fig:qualitative_applications}
\end{figure}

%% file: sec/10_conclusion.tex
\section{Conclusion}

We presented \method, a controllable constraint-conditioned dexterous grasp motion synthesis framework that translates heterogeneous task-driven specifications into physically stable grasping motions with a single policy.
Building on GraspXL, we extend its generalizable grasp generation with flexible controllability conditioned on optional and composable constraints. We organize constraints into four semantic levels and combine priority-aware composition with masked residual conditioning to support different constraint subsets. Dynamic hand centers
and feed-forward wrist guidance further improve control precision and grasp stability.

Extensive experiments show that \method provides accurate controllability under various individual and composed constraints, while preserving strong generalization across large-scale unseen objects and diverse hand morphologies.
It can serve as a plug-and-play low-level controller for different applications, such as whole-body grasp completion, functional grasping, and human-motion imitation.
We believe that bridging heterogeneous high-level task constraints with generalizable, contact-rich grasp control opens up a broader range of applications across many domains, including animation, robotics, and virtual reality.